%% file: main.tex
\documentclass[sigconf, screen,  nonacm]{acmart}
\AtBeginDocument{%
  }

\setcopyright{acmlicensed}
\copyrightyear{2018}
\acmYear{2018}
\acmDOI{XXXXXXX.XXXXXXX}
\acmConference[Conference acronym 'XX]{Make sure to enter the correct
  conference title from your rights confirmation email}{June 03--05,
  2018}{Woodstock, NY}
\acmISBN{978-1-4503-XXXX-X/2018/06}

\usepackage{fontawesome}

\usepackage{multirow} 

\usepackage{bbm}

\usepackage{algorithm}
\usepackage{algorithmic}
\usepackage{amsmath}
\usepackage{xcolor}
\usepackage{booktabs}
\usepackage{graphicx}

\usepackage{amssymb}
\usepackage{colortbl}
\usepackage{array}
\usepackage{makecell}
\usepackage{float}
\usepackage{pifont}
\usepackage{subcaption}

\usepackage{float}     % 图片浮动位置控制（如 [H] 选项）
\usepackage{caption}   % 更灵活的标题格式控制
\usepackage{subcaption} % 子图支持（如果有子图）

\usepackage{tabularx}

\usepackage{microtype}

\usepackage{tabularx}
\usepackage{multirow}

\usepackage{wrapfig}
\usepackage{booktabs}
\usepackage{booktabs} % 提供 \toprule, \midrule, \bottomrule
\usepackage{multirow}
\usepackage{geometry}
\usepackage[table]{xcolor} % 提供颜色支持
\definecolor{tablegray}{gray}{0.9} % 定义灰色，0.9 表示深浅，越接近1越白
\acmSubmissionID{965}

\begin{document}

\title{Beyond Chain-of-Thought: Rewrite as a Universal Interface for Generative Multimodal Embeddings}

\author{Peixi Wu$^{1}$, Ke Mei$^{1}$, Feipeng Ma, Bosong Chai$^{2}$, Zhibin Lan, Chenxi Zhao$^{2}$, \\
Shannan Yan$^{1}$, Jie Chen, Zhangchi Hu, Yansong Peng, Bo Lin$^{3}$, Junjie Zhou$^{1}$, \\
Dacheng Yin$^{1}$, Tianyi Wang$^{1}$, Fengyun Rao$^{1}$, Jing Lv$^{1}$, Hebei Li\textsuperscript{$\dagger$}, Xiaoyan Sun$^{4}$\\
$^{1}$WeChat Vision, Tencent Inc. $^{2}$Zhejiang University 
$^{3}$Tsinghua University \\
$^{4}$Institute of Artificial Intelligence, Hefei Comprehensive National Science Center\\
\texttt\tiny{\{peixiwu, raykoomei, feipengma, fengyunrao\}@tencent.com lihebei186@gmail.com} 
}
\thanks{\textsuperscript{$\dagger$} Corresponding Author}

\renewcommand{\shortauthors}{Peixi Wu, Ke Mei et al.}

\setcopyright{acmlicensed}
\copyrightyear{2026}
\acmYear{2026}
\acmDOI{XXXXXXX.XXXXXXX}
\acmConference[MM'26]{ACM Multimedia}{November 2026}{Rio de Janeiro, Brazil}

\settopmatter{printacmref=false}

\begin{abstract}
% 利用多模态大语言模型推进通用多模态嵌入已成为解决跨模态检索任务的关键方向。近期研究表明，相较于判别式方法，生成式思维链推理能显著提升嵌入表征质量。然而，思维链生成的推理轨迹往往过于冗长且总结回答容易存在语义粒度偏差，在需要查询自适应和语义对齐的通用检索场景中表现欠佳。为此，我们提出了一种以改写为核心的多模态嵌入框架，实现了生成和嵌入的联合优化和统一。本文核心贡献包括：（1）提出将文本改写作为通用接口，用自适应查询改写替代重度推理的思维链，跨多模态和多样化任务生成检索优化的嵌入表征；（2）设计交叉嵌入检索损失，桥接生成式与判别式嵌入的表征空间，来扩展生成式推理的灵活性，并提出细化强化学习，通过判别式嵌入的稳定监督引导生成式改写的优化；（3）在有限计算资源下，我们的框架在涵盖图像、视频和视觉文档的MMEB-V2基准测试的78项任务上，全面超越开创性的思维链生成式嵌入模型，同时生成长度缩短40%以上，推理效率显著提升。所提出的改写驱动范式展现出更强的零样本泛化能力与真实检索场景的适配性。本研究表明，针对性的查询改写而非开放式推理，是构建通用、高效且可扩展的生成式多模态嵌入系统的更优路径。

Multimodal Large Language Models (MLLMs) have emerged as a promising foundation for universal multimodal embeddings. 
Recent studies have shown that reasoning-driven generative multimodal embeddings can outperform discriminative embeddings on several embedding tasks.
However, Chain-of-Thought (CoT) reasoning tends to generate redundant thinking steps and introduce semantic ambiguity in the summarized answers in broader retrieval scenarios.
To address this limitation, we propose \textbf{R}ewrite-dr\textbf{I}ven \textbf{M}ultimodal \textbf{E}mbedding(\textbf{RIME}), 
a unified framework that jointly optimizes generation and embedding through a retrieval-friendly rewrite.
Meanwhile, we present the Cross-Mode Alignment (CMA) to bridge the generative and discriminative embedding spaces, 
enabling flexible mutual retrieval to trade off efficiency and accuracy.
Based on this, we also introduce Refine Reinforcement Learning (Refine-RL) that treats discriminative embeddings as stable semantic anchors to guide the rewrite optimization. 
% Refine RL描述不紧密
Extensive experiments on MMEB-V2, MRMR and UVRB demonstrate that 
RIME substantially outperforms prior generative embedding models while significantly reducing the length of thinking. 
\textit{Code is publicly available at \url{https://github.com/PeppaWu/RIME}.}
\end{abstract}



% caption不是很合理
% 准备放
% 1. UME-R1  2. Think-Then-Embed  3. GME 4. VLM2Vec 5. RIME
\begin{teaserfigure}
    \centering
    \includegraphics[width=\textwidth]{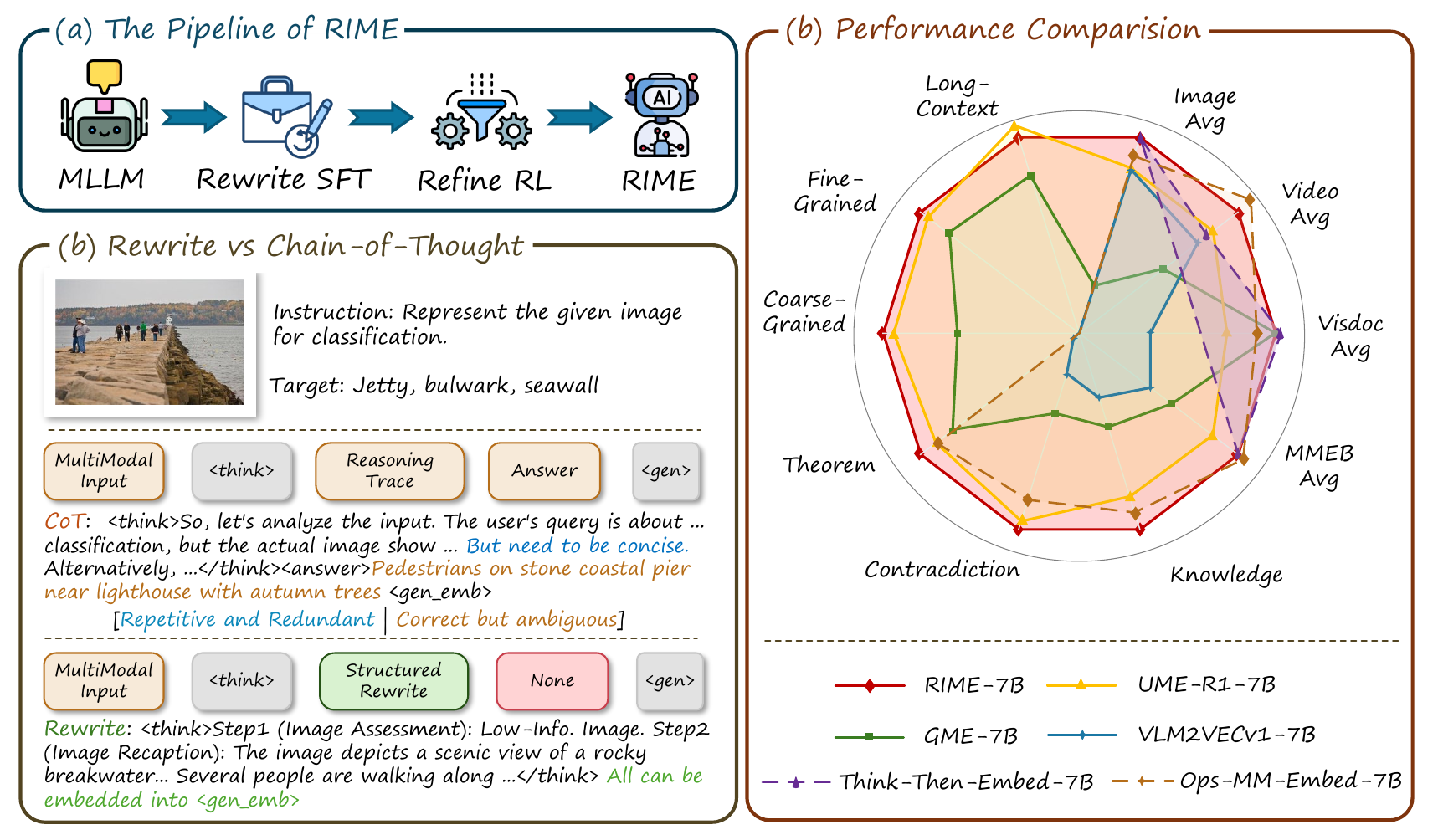}
    \caption{
An overview of the RIME pipeline. RIME utilizes rewrite SFT and refine RL to build generative embeddings, where rewrite offers a lightweight yet robust alternative to CoT. 
RIME achieves superior  performance across multiple benchmarks.
    }
    \label{fig:head}
\end{teaserfigure}


\maketitle

\input{papers/1_Introduction}

\input{papers/2_Related_Work}

\input{papers/3_Method}

\input{papers/4_Experiment}

\input{papers/5_Conclusion}

\bibliographystyle{ACM-Reference-Format}
\bibliography{papers/bibfile}

\input{appendix/appendix}
%\input{papers/x_app}

\end{document}

%% file: papers/1_Introduction.tex
\section{Introduction}
% 多模态嵌入学习领域在多模态大型语言模型（MLLM）的快速发展推动下发生了变革性转变。与传统双编码器架构（如 CLIP、BLIP）不同，这类架构在处理复杂多模态输入和结构化视觉数据（如视频和多页文档）时面临挑战，而基于 MLLM 的嵌入模型（如 VLM2Vec、GME、MM-Embed）已成为通用解决方案。通过融合指令微调与对比学习，这些模型擅长将异构数据（包括文本、图像、视频和视觉文档）编码到统一的稠密表示空间中。这些基于 MLLM 的模型的核心优势在于，它们能够通过固有的指令跟随能力实现跨模态语义对齐，标志着从模态特定编码向统一多模态表示学习的范式转变。 

The field of multimodal embedding learning has undergone a transformative shift with the rapid development of Multimodal Large Language Models (MLLMs). 
Traditional dual-encoder architectures such as CLIP~\cite{radford2021learning} and BLIP~\cite{li2023blip} struggle with complex multimodal inputs and structured visual data like visual documents and videos. 
In contrast, MLLM-based embedding models such as VLM2Vec~\cite{jiang2024vlm2vec,zhai2023sigmoid,jia2021scaling}, GME~\cite{zhang2024gme}, and MM-Embed~\cite{lin2024mm} have emerged as general solutions. 
By integrating instruction fine-tuning with contrastive learning, these models~\cite{jiang2024e5,lan2025llave,jiang2024vlm2vec,zhang2024gme,lan2025ume,liu2025lamra,gu2025breaking,kong2025modality,zhou2025megapairs} excel at encoding heterogeneous data including text, images, videos and visual documents into a unified dense representation space. 
Their core advantage lies in achieving multimodal semantic alignment through inherent instruction-following capabilities built upon powerful MLLM backbones~\cite{liu2023visual,li2024llava,li2024llava_next,wang2024qwen2,bai2025qwen2,Qwen3_VL}, marking a paradigm shift from modality-specific encoding to unified multimodal representation learning.

% 紧随判别式嵌入的发展，生成式多模态嵌入正逐渐成为一个备受关注的前沿方向，其核心目标在于充分释放多模态大语言模型在生成与推理方面的潜能。与直接从输入 token 中提取嵌入的判别式方法不同，生成式嵌入通过引入中间生成环节来增强语义理解的深度。在这一方向上，思维链推理成为主流范式，它引导模型在生成嵌入之前形成分步的逻辑推导，并最终通过输出答案来实现对检索任务的前置引导。然而，在实际应用中，部分前期研究发现，对比学习与自回归生成联合训练存在内在的不兼容性——同时优化生成目标与嵌入对比目标往往会削弱嵌入的判别能力。
In the wake of the development of discriminative embeddings, generative embeddings are gradually emerging as a frontier area attracting substantial attention~\cite{lan2025ume,cui2025think,jiang2026embed}. The core objective here is to comprehensively unleash the potential of multimodal large language models in terms of generation and reasoning. Unlike discriminative embedding models that directly extract embeddings from a specific token, generative embeddings introduce an intermediate reasoning process to enhance semantic understanding.
Within this paradigm, chain-of-thought (CoT)~\cite{wei2022chain} has become the dominant approach. It guides the model to form step-by-step logical deductions before generating embeddings, guiding retrieval tasks via a summarized answer. However, recent studies~\cite{cui2025think,yu2025cafe,chen2025moca} have revealed an inherent incompatibility between joint training of contrastive learning and autoregressive generation. In other words, optimizing both objectives simultaneously often weakens the discriminative capabilities of generative embeddings.

% 为解决这一问题，研究人员提出了“先思考后嵌入”（Think-then-Embed）的解耦范式，将思维链生成与嵌入编码过程分离。尽管该设计在一定程度上缓解了训练目标的冲突，却也带来了模型参数的大量冗余。此外，由于检索任务在不同场景下对答案粒度的要求各异，思维链微调所获得的推理器与嵌入器之间往往难以在未见过的场景下实现良好兼容。与此同时，另一部分工作（如 UME-R1）尝试保留联合训练框架，借助基于可验证奖励的强化学习来缓解生成与嵌入之间的目标冲突。虽然该方法在促进二者对齐方面取得了一定进展，但其引入的长思维链过程增加了计算开销，难以满足大规模通用检索系统对低延迟和高吞吐的实际需求。更重要的是，无论采用解耦式还是耦合式架构，基于思维链的生成范式在通用检索场景中都暴露出一系列固有局限：
% (1) 逻辑推导的分步设计虽适合复杂推理，却在简单语义检索中引入大量冗余；而面向全局语义生成的总结式答案，又难以支持细粒度检索任务的需求。(2) 生成冗长推理链所引发的高计算成本，在大规模低延迟检索系统中难以承受。上述缺陷揭示了当前推理驱动型生成范式与现实世界中多模态检索对通用性、效率与可扩展性的核心需求之间的根本性错位。
To address this issue, researchers propose the Think-then-Embed (TTE) decoupling paradigm~\cite{cui2025think,cui2025reason,jiang2026embed}, which separates CoT generation from embedding encoding using an external reasoner but introduces substantial parameter redundancy. Meanwhile, other works~\cite{lan2025ume,liu2025reasoning,zhu2025retrv} retain joint training with reinforcement learning from verifiable rewards to optimize CoT and embedding together, yet the long CoT increases inference cost, hindering low-latency, high-throughput deployment. 
More importantly, regardless of decoupled or coupled architectures, the CoT-based paradigm faces a series of inherent limitations: 
% 适当举例说明（1） Long Doc类比过来，给一点具体的解释，无意义的反思过程
(1) \textbf{Reasoning redundancy}: CoT generates many intermediate steps that are often unnecessary in general retrieval. 
(2) \textbf{Target Ambiguity}: In general retrieval, targets are multi-granular, non-unique, and ambiguous. The specific summarized answer from CoT can easily mislead embeddings or cause semantic bias. (3) \textbf{Inference Inefficiency}: Generating long CoT incurs high inference costs, which is unbearable in practical retrieval systems. These defects reveal that the CoT-based paradigm fundamentally misaligns with the generality, efficiency, and scalability required in real-world multimodal retrieval.

% 为突破上述局限，我们提出了一种全新的生成式多模态嵌入通用范式，其核心在于以检索优化的重写机制超越传统的思维链推理。该机制不再生成推理轨迹和总结式答案，而是通过重写将多模态输入重构为统一的、检索优化的语义表达，不强制产生复杂的思维链来推理出一个最终的答案而是对多模态的输入进行解释或者扩展，并确保关键信息的完整性。这一重写驱动的框架具备模态无关性与任务通用性，可无缝适配多粒度检索，复杂逻辑检索，多种模态混合检索等场景。我们在一个统一模型中实例化了该范式，该模型能够同时生成判别式嵌入与基于重写的生成式嵌入，从而灵活应对多样化的应用需求。

To overcome the above limitations, we propose a novel paradigm for generative embeddings. Its core is a retrieval-friendly rewriting mechanism that goes beyond traditional CoT reasoning. Instead of generating reasoning trajectories and summarized answers, it only structurally rewrites the images, videos and documents for recaption and texts for further explanation without summarizing a final answer. This rewrite-driven framework is modality-agnostic and task-general, and can seamlessly adapt to scenarios such as multi-granularity retrieval, complex logical retrieval, and mixed-modality retrieval. We instantiate this paradigm in a unified model that can simultaneously generate discriminative embeddings and rewrite-based generative embeddings.

% 为了缓解推理重写过程中产生的推理成本，我们提出了交叉嵌入检索损失，旨在对齐判别式嵌入与生成式嵌入所表征的语义空间。该损失通过对比学习，交叉拉近同一查询下正样本的两种嵌入，同时推远负样本，从而初步确保生成式嵌入能够继承判别式嵌入的判别能力，避免生成过程引入无关噪声。这一机制也赋予了判别式嵌入与生成式嵌入之间的互相检索能力，显著提升了生成式嵌入的灵活性与适用性。在此基础上，由于判别式与生成式的向量空间基本可比，生成式向量空间可视为判别式空间的细化与增强（refine）。因此，我们进一步引入了基于双空间对齐的强化学习范式。该范式将判别式嵌入作为稳定的语义锚点，以其检索性能作为奖励信号，优先鼓励那些在重写过程中生成的、其向量信息量高于判别式嵌入的样本，从而引导模型朝向语义更丰富、判别能力更强的方向优化。
To mitigate the inference overhead incurred by the rewriting process, we propose the cross-mode alignment (CMA) mechanism. Via mutual contrastive learning, the mechanism aligns discriminative and generative embeddings for identical queries, while separating those of negative samples. It facilitates cross-retrieval between the two embedding spaces, thereby enhancing the flexibility and applicability of the generative embeddings.
Based on CMA, the embedding spaces of discriminative and generative embeddings are basically aligned. The generative embedding can be viewed as a refinement of the discriminative embedding. Therefore, we further introduce a refine reinforcement learning method (Refine-RL).
By taking discriminative embeddings as stable semantic anchors and using whether the similarity gap of generative embeddings exceeds that of discriminative embeddings as a process reward, we encourage a more effective rewriting process. This guides the model toward optimized semantic richness and stronger discriminative capability.

% 实验表明，我们的模型在测试推理长度相对之前模型降低至少40\%的情况下,在MMEB-V2上面远远超过之前的工作，并通过消融实验验证了不同粒度下rewrite范式会比思维链对于检索场景更具备鲁棒性和泛化性。总而言之，我们的工作的主要共享点可以被总结为以下几点：

%
Experiments show that our model reduces thinking process by approximately \textbf{50\%} compared with prior methods~\cite{lan2025ume,liu2025reasoning}, while significantly outperforming existing works on MMEB-V2~\cite{jiang2024vlm2vec}. In summary, our main contributions are as follows:
% \begin{itemize}
%     \item  提出了一种以重写机制为核心的生成式多模态嵌入新范式，取代传统的思维链推理机制，通过重构输入为统一的语义表达，实现模态无关、任务通用的高效检索接口。
%     \item 设计了交叉嵌入检索损失，利用对比学习对齐判别式嵌入与生成式嵌入的语义空间，使生成式嵌入继承判别能力，同时支持两者间的互相检索，显著提升嵌入的灵活性与适用性。
%     \item 引入了基于双空间对齐的强化学习范式，以判别式嵌入为稳定的语义锚点，以其检索性能作为奖励信号，引导生成式嵌入在重写过程中保留更丰富的信息，实现语义空间的细化与增强。
%     \item 该方法在测试推理长度平均降低至少40\%的情况下，在MMEB-V2，UVRB和MR2Bench基准上大幅超越现有方法，并通过消融实验验证了重写范式相较于思维链在多粒度检索场景中的鲁棒性与泛化性。
% \end{itemize}

\begin{itemize}
\item  We propose Rewrite-Driven Joint SFT as a novel universal paradigm for generative embeddings, replacing the traditional CoT reasoning.
\item  We design the cross-mode alignment mechanism that aligns discriminative and generative embeddings, enabling mutual retrieval and improved flexibility.
\item  We introduce the refine reinforcement learning that uses discriminative embeddings as anchors, guiding the rewriting process to better refine the generative embedding.
% MRMR单独强调一下
\item  Extensive experiments demonstrate that RIME significantly outperforms existing generative embedding models on the MMEB-V2~\cite{jiang2024vlm2vec}, UVRB~\cite{guo2025towards}, and MRMR~\cite{zhang2025mrmr} benchmarks.
    % \item 设计了交叉嵌入检索损失，利用对比学习对齐判别式嵌入与生成式嵌入的语义空间，使生成式嵌入继承判别能力，同时支持两者间的互相检索，显著提升嵌入的灵活性与适用性。
    % \item 引入了基于双空间对齐的强化学习范式，以判别式嵌入为稳定的语义锚点，以其检索性能作为奖励信号，引导生成式嵌入在重写过程中保留更丰富的信息，实现语义空间的细化与增强。
    % \item 该方法在测试推理长度平均降低至少40\%的情况下，在MMEB-V2，UVRB和MR2Bench基准上大幅超越现有方法，并通过消融实验验证了重写范式相较于思维链在多粒度检索场景中的鲁棒性与泛化性。
\end{itemize}

%% file: papers/2_Related_Work.tex
\section{Related Works}

\subsection{Multimodal Representation Learning}
% 多模态嵌入学习旨在将来自不同模态的信息（如文本、图像、视频及文档）映射到统一的语义表示空间，以支持跨模态检索任务。早期研究主要依赖双编码器（dual-encoder）架构，通过对比学习实现跨模态对齐。其中，CLIP 等方法通过大规模图文对训练，实现了图像与文本之间的共享嵌入空间，并在跨模态检索任务中取得显著成功。随后，BLIP、ALIGN 等工作进一步结合视觉编码器与语言模型，提高了多模态语义建模能力。然而，这类方法通常依赖固定结构的模态编码器，对复杂输入结构（如长视频、多页文档或多模态混合输入）的建模能力仍然有限。
% 随着多模态大型语言模型（MLLM）的快速发展，研究者开始探索基于生成式架构构建统一的嵌入模型。与传统双编码器方法不同，这类模型利用大型语言模型强大的语义理解能力，通过指令微调和对比学习将不同模态数据映射到统一表示空间。例如，VLM2Vec、GME 以及 MM-Embed 等方法通过共享的语言建模框架，实现了图像、文本及视觉文档等多模态数据的统一编码。这类方法的核心优势在于其天然具备的指令理解能力，使模型能够在复杂查询条件下实现更加灵活的跨模态语义对齐。

Multimodal representation learning aims to embed heterogeneous modalities into a unified semantic space for multimodal retrieval. Early works mainly adopt dual-encoder architectures trained with contrastive learning~\cite{zhai2023sigmoid,radford2021learning,oord2018representation}. For example, CLIP~\cite{radford2021learning} learns shared image-text representations from large-scale web data and achieves prominent cross-modal retrieval performance. Follow-up models including BLIP~\cite{li2023blip} and ALIGN~\cite{jia2021scaling} further combine vision encoders with language models to strengthen semantic alignment. Nevertheless, these methods adopt fixed-structure encoders and lack sufficient modeling capability for complex inputs such as videos, multi-page documents and mixed-modal content.

The emergence of multimodal large language models~\cite{liu2023visual,li2024llava,li2024llava_next,wang2024qwen2,bai2025qwen2,Qwen3_VL} has opened a new paradigm for unified embedding learning. Unlike traditional dual-encoder structures, MLLMs leverage strong semantic understanding to project heterogeneous modalities into a shared space via instruction tuning and contrastive learning. Typical models including VLM2Vec~\cite{jiang2024vlm2vec,meng2025vlm2vec}, GME~\cite{zhang2024gme}, MM-Embed~\cite{lin2024mm}, and LLaVE~\cite{lan2025llave} adopt unified LLM backbones for multimodal inputs. Subsequent studies~\cite{liu2025lamra,gu2025breaking,kong2025modality,wu2026tsembed,jiang2026embed} further boost the scalability and performance of MLLM-based embeddings. A core advantage lies in inherent instruction-following capability, enabling flexible multimodal alignment in complex queries.

% 重复语义标题
\subsection{ Multimodal Generative Embedding }
% 在传统判别式嵌入模型的基础上，生成式嵌入逐渐成为近年来的重要研究方向。该类方法利用生成模型的推理能力，在生成过程中对输入语义进行显式建模，从而提升嵌入表示的语义深度。一个典型思路是引入中间生成过程，通过生成解释、答案或推理步骤来辅助嵌入学习。在这一范式中，思维链（Chain-of-Thought, CoT）推理被广泛用于增强模型对复杂查询的理解能力。相关研究通常引导模型在生成嵌入之前先生成分步推理过程，再通过最终答案和推理轨迹提取语义表示，从而在复杂推理任务中获得更具结构化的语义表达。
% 然而，后续研究发现，生成式训练目标与嵌入学习中的对比目标之间可能存在冲突。部分工作指出，在联合优化自回归生成目标与嵌入对比学习目标时，模型往往更倾向于优化语言生成质量，从而削弱嵌入的判别能力。为缓解这一问题，一些研究提出“先思考后嵌入”（Think-then-Embed）的解耦框架，将推理生成模块与嵌入编码模块分离，以避免训练目标冲突。另一类方法则尝试在联合训练框架下，通过强化学习或可验证奖励机制来协调生成与嵌入目标。另外，尽管这些方法在一定程度上提升了生成式嵌入的效果，但其依赖长推理链的生成过程往往带来较高的计算成本，并可能在简单检索场景中引入不必要的冗余推理。
Generative embedding has become a major research direction beyond traditional discriminative models.By exploiting the reasoning ability of generative models, these methods explicitly model input semantics during generation to strengthen representation quality.A typical paradigm introduces intermediate generative steps (e.g., reasoning chains) to facilitate embedding learning.In this line, Chain-of-Thought (CoT)~\cite{wei2022chain,guo2025deepseek,wang2025multimodal,kojima2022large} is widely used to enhance complex query understanding.Most models~\cite{jiang2026embed,liu2025rematch,liu2025idmr,lan2025ume} first generate stepwise rationales and then output final embeddings.

However, recent studies~\cite{yu2025cafe,chen2025moca} reveal inherent conflicts between generative objectives and contrastive learning in embedding models. Joint optimization of autoregressive generation and embedding alignment often biases models toward language modeling, impairing embedding discriminability~\cite{yu2025cafe,liang2026learn, wu2025spiking, wu2025efficient}. To alleviate this issue, decoupled designs such as Think-then-Embed (TTE)~\cite{cui2025think,cui2025reason} separate reasoning generation from embedding encoding. Other works~\cite{lan2025ume} adopt reinforcement learning and reward mechanisms to balance two goals in one framework. Nevertheless, the CoT-based methods introduce substantial computational overhead and may yield redundant thinking.

\subsection{Retrieval-oriented Query Rewrite}
% 查询重写作为信息检索领域的长期研究热点，其核心目标在于弥合用户查询与文档内容之间的语义鸿沟。早期研究主要集中于词汇层面的重构技术，如基于伪相关反馈的查询扩展与关键词重加权，旨在提升稀疏查询场景下的召回性能。近年来，大语言模型（Large Language Models, LLMs）的兴起推动了查询重写向语义层面的演进，使其能够分解复杂指令、推断隐含约束并生成多样化的候选查询。

% 然而，现有的生成式查询嵌入方法多依赖于思维链（Chain-of-Thought, CoT）推理机制，其冗长的分步推导与答案总结过程引入了较强的任务特定性，难以适用于通用的检索场景。针对这一问题，我们提出了一种面向检索优化的查询改写机制，摒弃冗余推理步骤，直接将原始查询重构为检索友好的语义表达形式，不强制生成总结性答案。该机制兼具轻量级与自适应特点，在保障信息完整性的同时显著降低了推理开销。
% 值得注意的是，已有研究将查询重写引入复杂推理检索任务并取得显著成效，但相关方法仍以离线方式运行，且大多局限于单一模态（如仅对图像查询进行改写），忽视了对文本等其他模态的扩展与语义重构。

Query rewrite aims to bridge the semantic gap between user queries and documents~\cite{carpineto2012survey}. Early lexical methods include pseudo-relevance feedback~\cite{xu2009query,li2023pseudo,wang2023colbert} and keyword reweighting~\cite{aklouche2023discriminative} to enhance recall for sparse queries.Recently, text-based query rewriting has advanced greatly~\cite{ma2023query}, mainly using LLMs for query expansion, rationale generation, or pseudo-document construction. MLLMs further enable semantic-level rewriting, supporting query decomposition, implicit constraint inference, and diverse query generation.While prior work~\cite{cui2025reason,zhou2025mr,zhang2025mrmr} applies query rewrite in complex reasoning retrieval, it often relies on external LLMs for pre-retrieval expansion or reasoning and is modality-specific~\cite{uzan2025guided,lin2026ark}, focusing on images and ignoring other modalities like text.Additionally, existing generative multimodal embedding models~\cite{lan2025ume,jiang2026embed,cui2025think} rely heavily on CoT, 
whose stepwise derivation and summarization introduce task biases and limit general retrieval. Unlike CoT, RIME uses the rewrite paradigm to extend multimodal information, supporting interpretation and expansion across all modalities.
\begin{figure}[!t]
    \centering
    \includegraphics[width=\columnwidth]{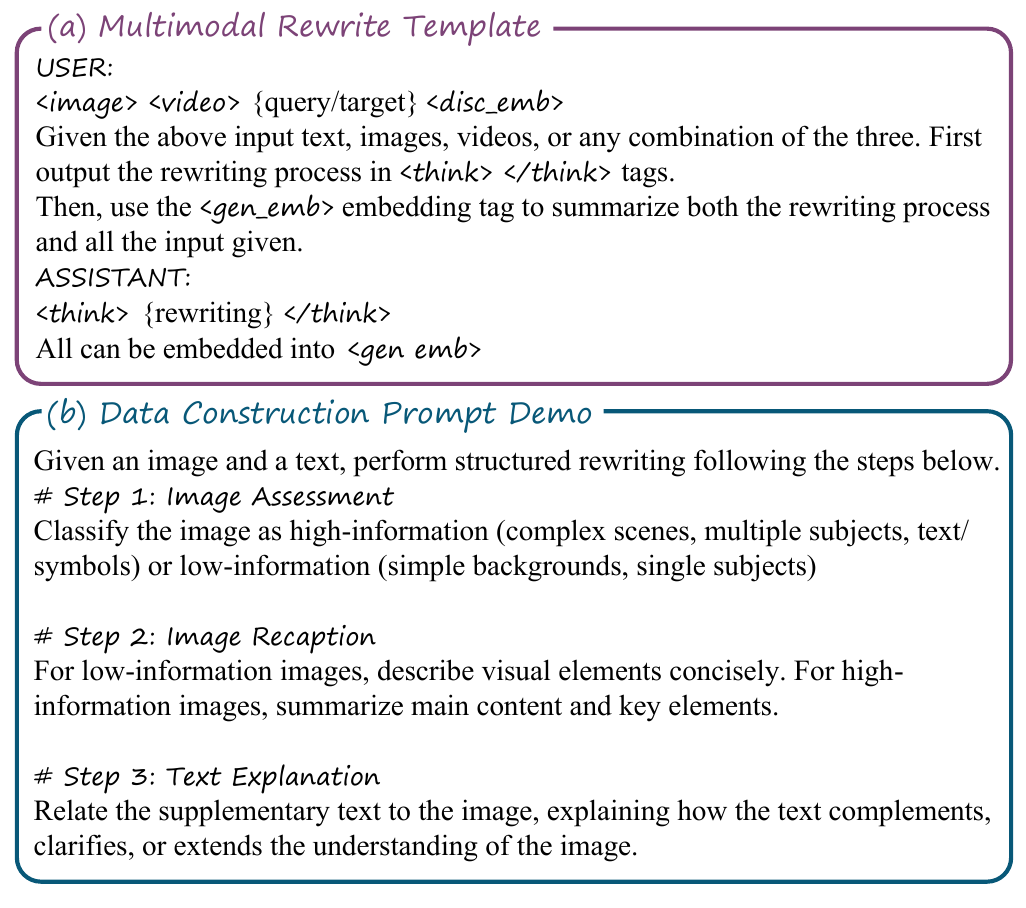}
    \caption{
    The complete multimodal rewrite template and the prompt demo used for SFT cold-start data construction.
    }
    \label{fig:temp}
\end{figure}

%% file: papers/3_Method.tex
\section{Method}

This section begins with the preliminary definitions, including
the background of discriminative and generative embedding (Section 3.1). We then introduce our proposed rewrite-driven
joint supervised fine-tuning framework that unifies multimodal
contrastive learning with retrieval-oriented rewrite generation (Section 3.2).
Next, we introduce cross-mode alignment to bridge discriminative and generative embedding spaces (Section 3.3). Finally, we
propose the refine reinforcement learning  that guides rewrite policy optimization through stable
supervision from discriminative embeddings (Section 3.4).

% B 部分可以换成一个表征空间（比如一个圆）
\begin{figure*}[!t]
    \centering
    \includegraphics[width=\textwidth]{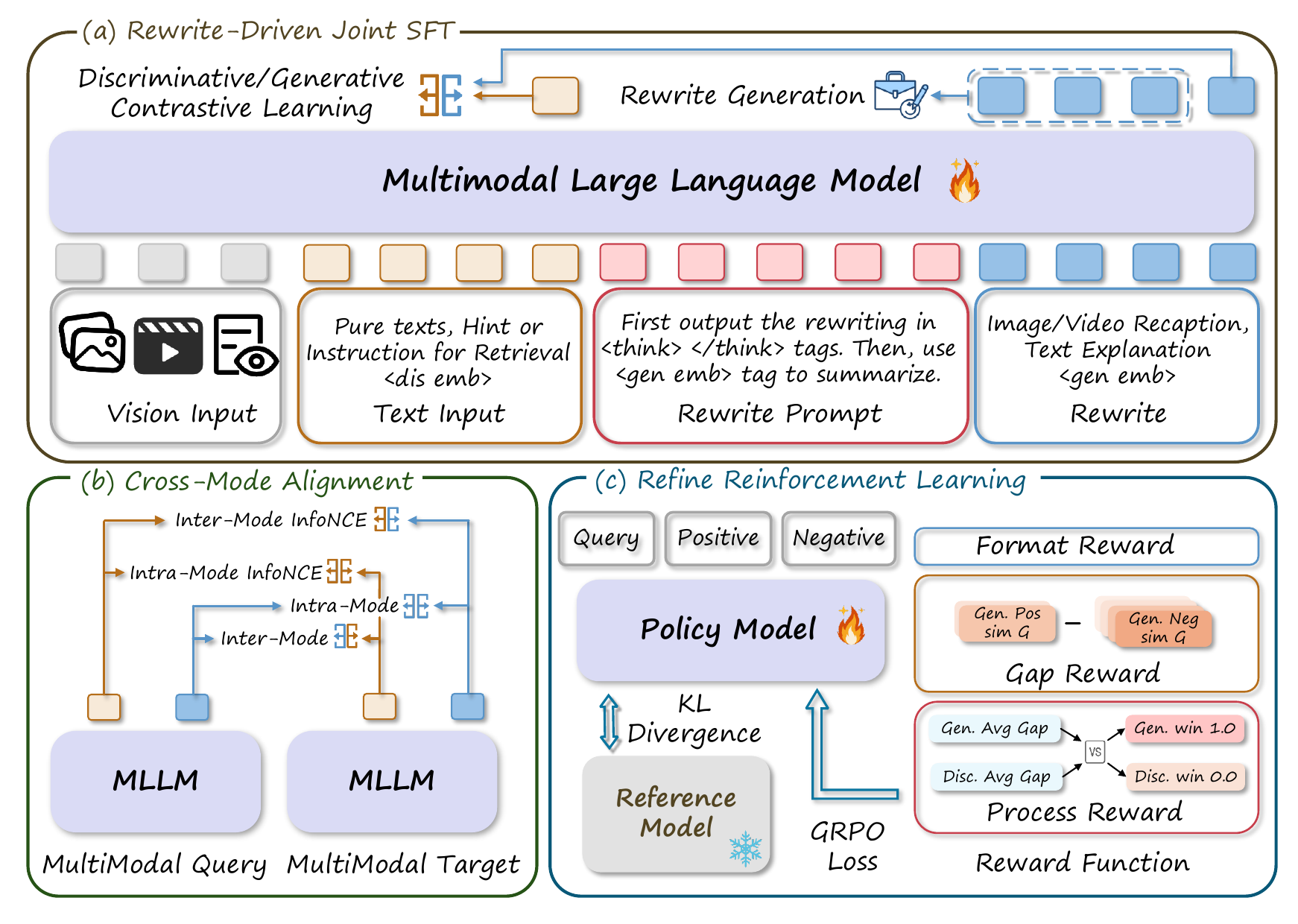}
    \caption{
Overview of RIME framework. The framework comprises three core components: (a) Rewrite-Driven Joint SFT that unifies contrastive learning with rewrite generation; (b) Cross-Mode Alignment via inter-mode and intra-mode InfoNCE losses; (c) Refine Reinforcement Learning that guides rewrite policy optimization with multiple rewards.
    }
    \label{fig:main}
\end{figure*}

\subsection{Preliminary}
% 我们首先简要介绍判别式与生成式嵌入模型的技术背景。对于这两种范式，其目标都是在给定查询$q$的情况下，从候选目标集$\mathcal{T}$中检索最相关的目标$t^+$。对于判别式嵌入模型，它们直接编码多模态输入，并提取最后一个输入标记的归一化表示作为判别式嵌入，通过InfoNCE损失进行优化：

% \begin{equation}
% \mathcal{L}{disc} = -\frac{1}{N} \sum{i=1}^{N} \log \frac{\exp(f_\theta(q_i) \cdot f_\theta(t_i) / \tau)}{\sum_{j=1}^{N} \exp(f_\theta(q_i) \cdot f_\theta(t_j) / \tau)}
% \end{equation}

% 其中$\tau$为温度超参数。与判别式方法不同，生成式嵌入模型在生成最终表示之前，首先产生中间的推理轨迹和摘要。模型遵循特定模板生成思维链<think>{reasoning}</think>和摘要<answer>{summary}</answer>，最终使用特定标记的隐藏状态作为生成式嵌入。其优化过程可形式化为：

% \begin{equation}
% o^{q_i} = f_\phi(q_i), \quad o^{t_j} = f_\phi(t_j)
% \end{equation}

% \begin{equation}
% \mathcal{L}{gen} = -\frac{1}{N} \sum{i=1}^{N} \log \frac{\exp(f_\theta(q_i, o^{q_i}) \cdot f_\theta(t_i, o^{t_i}) / \tau)}{\sum_{j=1}^{N} \exp(f_\theta(q_i, o^{q_i}) \cdot f_\theta(t_j, o^{t_j}) / \tau)}
% \end{equation}

% 这里，$f_\phi$可以是外部推理模型，也可以是与$f_{\theta}$共享参数的模型，这两种方式代表了生成式嵌入模型的两种不同技术路径。考虑到外部推理模型存在的参数冗余和优化兼容性问题，本文采用后者作为基线。
We begin by briefly introducing the technical background of discriminative and generative embedding models. For both paradigms, the objective is to retrieve the most relevant target 
$t^+$ from a set of candidate targets $\mathcal{T}$ given a specific query $q$. For discriminative embedding models, they directly encode the multimodal inputs and extract the normalized representation of the last input token as the discriminative embedding, optimized via the InfoNCE loss:
\begin{equation}
\mathcal{L}_{disc} = -\frac{1}{N} \sum_{i=1}^{N} \log \frac{\exp(f_\theta(q_i) \cdot f_\theta(t_i) / \tau)}{\sum_{j=1}^{N} \exp(f_\theta(q_i) \cdot f_\theta(t_j) / \tau)}
\end{equation}
where $f_\theta$ denotes the embedding model with final normalization, and $\tau$ represents the temperature hyper-parameter. In contrast, generative embedding models first produce intermediate reasoning trajectory and summarization before generating the final representation. The model follows a specific template to generate the chain-of-thought <think>{reasoning}</think> and summary <answer>{summary}</answer>, and finally uses the hidden state of a specific token as the generative embedding. The optimization process can be formulated as:
\begin{equation}
o^{q_i} = f_\phi(q_i), \quad o^{t_j} = f_\phi(t_j)
\end{equation}
\begin{equation}
\mathcal{L}_{gen} = -\frac{1}{N} \sum_{i=1}^{N} \log \frac{\exp(f_\theta(q_i, o^{q_i}) \cdot f_\theta(t_i, o^{t_i}) / \tau)}{\sum_{j=1}^{N} \exp(f_\theta(q_i, o^{q_i}) \cdot f_\theta(t_j, o^{t_j}) / \tau)}
\end{equation}
Here, $o$ denotes the generated reasoning process, and $f_\phi$ can be either an external model or share parameters with $f_{\theta}$, representing two distinct approaches for generative embedding models. Given the parameter redundancy and optimization issues of external reasoning models, we adopt the latter as the baseline.

\subsection{Rewrite-Driven Joint SFT}
% 这一段主要是要说明 Rewrite 和 CoT在数学上的区别，训练过程也需要说明，主要的是CoT的冗余性.

% 生成式嵌入模型当前普遍借鉴大语言模型中的思维链范式，却忽视了该范式在嵌入任务中的适用性边界。思维链的初衷在于通过显式推理引导模型生成确定性答案，如在数学推理中获取精确结果。然而，嵌入任务的目标特性与此存在本质差异：检索查询往往不具备唯一解，甚至检索目标本身就是陈述而不存在answer。在此情境下，直接将思维链逻辑迁移至嵌入生成，易导致总结粒度与检索需求错位，引入语义偏差。进一步地，尽管推理密集型检索场景对思维链具有一定依赖性，但此类场景呈现显著的任务特异性。面向通用检索任务时，思维链过程往往诱发冗余思考甚至错误推理，带来较高的学习难度和噪声。

% 基于上述观察，本文提出\textbf{重写驱动的联合监督微调框架}。该框架回归检索友好的重写范式：摒弃强制模型生成思维链式推理与答案粒度总结的策略，转而采用重写驱动方式，使模型学习对图像、视频及文本内容进行描述性重构，
% 这样的重写方式依旧采用结构化的思考步骤，但仅仅将其作为分治和不同场景下描述的细化要求，本质上并不是推理而是重写。
% 相较于传统查询重写方法，本框架在两个维度上实现扩展：其一，\textbf{重写对象的对称性}——不仅对查询侧进行重写，同时对目标侧进行重写，以确保查询与目标在重写语义空间中的精准对齐；其二，\textbf{优化目标的联合性}——摒弃离线式外部重写模型范式，使嵌入模型在端到端训练中联合优化重写生成与对比学习。重写生成通过自回归语言建模损失优化，保障生成内容的语义保真度；同时，基于重写后的查询与目标表示，通过对比学习损失拉近正样本对、推开负样本对，使重写过程直接服务于检索性能提升。联合损失函数形式化如下：

% \begin{equation}
% \mathcal{L}_{\text{joint}} = \mathcal{L}_{\text{CM-InfoNCE}} + \lambda \cdot \mathcal{L}_{\text{Rewrite}}
% \end{equation}

% 其中，\(\mathcal{L}_{\text{CM-InfoNCE}}\)为交叉模式的对比损失；\(\mathcal{L}_{\text{Rewrite}}\)为重写生成的自回归语言建模损失；超参数\(\lambda\)平衡两项损失。通过该联合优化目标，模型在生成检索友好重写内容的同时，将重写语义直接融入嵌入表示。

Most current generative embedding models adopt CoT paradigms from LLMs, yet overlook their applicability boundaries in embedding tasks.
CoT is originally designed to guide models toward deterministic answers through explicit reasoning, such as obtaining solutions in mathematical reasoning. However, the goal of embedding tasks is fundamentally different: retrieval queries often do not have unique answers or summarizations, and retrieval targets are usually statements which do not need to be further summarized.
Directly applying CoT to embedding generation can easily cause mismatches between summarization granularity and retrieval needs, introducing semantic distortions.
Furthermore, although reasoning-intensive retrieval scenarios rely on CoT, these scenarios are highly task-specific.
For general retrieval tasks, CoT often leads to redundant reasoning or even erroneous inference, increasing training difficulty and introducing unnecessary noise.
%% Rewrite 强调一下结构化的本质

Based on the above insights, we propose a rewrite-driven jointly supervised fine-tuning framework. As shown in Figure \ref{fig:temp} and \ref{fig:main}, the framework presents a retrieval-friendly rewriting paradigm: instead of forcing the model to generate CoT reasoning and summarized answers, it adopts a rewrite-driven strategy that learns descriptive reconstruction of images, videos, and text. 
Specifically, this rewriting process uses structured steps as a divide-and-conquer strategy for refining descriptions across scenarios as shown in Figure~\ref{fig:temp}, prioritizing rewriting over reasoning.
Furthermore, compared with traditional query rewriting methods, we rewrite not only the query side but also the target side, ensuring alignment between queries and targets in the rewritten semantic space. Also, we abandon the offline external rewriting paradigm and jointly optimize rewrite generation and contrastive learning in an end-to-end training pipeline. The joint loss is formalized as follows:
\begin{equation}
\mathcal{L}_{\text{Joint}} = \lambda \cdot \mathcal{L}_{\text{Rewrite}} + \mathcal{L}_{\text{CM\_InfoNCE}}
\end{equation}
where $\mathcal{L}_{\text{Rewrite}}$ denotes the autoregressive language modeling loss. $\mathcal{L}_{\text{CM\_InfoNCE}}$ is the cross-mode contrastive loss. The hyperparameter $\lambda$ balances the trade-off between rewrite quality and retrieval performance. The joint loss enables end-to-end learning of semantically consistent and retrieval-effective multimodal embeddings.

% \begin{figure}[!t]
%     \centering
%     \includegraphics[width=\columnwidth]{figures/Chattemp.pdf}
%     \caption{
%     The complete multimodal rewrite template.
%     }
%     \label{fig:temp}
% \end{figure}

\subsection{Cross-Mode Alignment}
% 这一段要强调rewrite过程仍不可避免的产生推理时间的消耗，对于嵌入模型的实时性带来了考验，所以我们需要crosss-mode的对齐，一是可以让检索的时候灵活选取mode，缓解生成式嵌入的高推理成本，二是初步确保生成式嵌入能够继承判别式嵌入的判别能力，减少生成过程引入无关噪声。

%尽管重写驱动范式有效缓解了思维链的冗余问题，但其生成过程仍不可避免地引入额外的推理时计算开销，这对嵌入模型的实时性提出了严峻挑战。重写生成的自回归解码过程显著增加了查询处理延迟，难以满足大规模检索场景对低延迟的苛刻要求。
% 为应对这一挑战，我们引入跨模态对齐机制，通过跨嵌入检索损失实现生成式与判别式嵌入之间的表示对齐。该机制带来双重收益：其一，推理时模态灵活切换——通过对齐两种嵌入的表示空间，模型在推理阶段可根据实时性需求灵活选择使用判别式嵌入或生成式嵌入，在保证检索性能的前提下有效减少生成过程的推理开销；其二，判别能力继承与噪声抑制——对齐目标迫使生成式嵌入在语义空间中向判别式嵌入靠拢，从而继承其强判别能力，同时抑制重写生成过程中可能引入的任务无关噪声，确保生成式嵌入的语义纯度。
% 具体来说，如图所示，跨模态对齐机制，以跨嵌入检索损失作为对齐信号，在原来模式内对齐的基础上，要求生成式查询嵌入能够有效检索判别式目标嵌入，反之亦然，从而强制执行对齐。交叉模式检索损失形式化如下：

Although rewrite-driven paradigms effectively alleviate redundancy in chain-of-thought reasoning, their generation inevitably introduces extra inference-time computation. The autoregressive decoding in rewriting significantly increases query processing latency, making it difficult to meet the strict low-latency requirements of large-scale retrieval scenarios.

To address this challenge, we introduce a cross-mode alignment (CMA) mechanism that aligns representations between generative and discriminative embeddings via cross-mode InfoNCE loss. By encouraging mutual retrieval between the two embedding spaces, it enables flexible inference pathways: during deployment, the system can choose between directly computing discriminative embeddings for low-latency retrieval or invoking generative rewriting when deeper semantic reasoning is required. Together, cross-mode alignment can be  formulated as:
\begin{equation}
\begin{split}
\mathcal{L}_{intra} = & -\frac{1}{N} \sum_{i=1}^{N} 
\log \frac{\exp(f_\theta(q_i) \cdot f_\theta(t_i, o^{t_i}) / \tau)}
{\sum_{j=1}^{N} \exp(f_\theta(q_i) \cdot f_\theta(t_j, o^{t_j}) / \tau)} \\
& -\frac{1}{N} \sum_{i=1}^{N} 
\log \frac{\exp(f_\theta(q_i, o^{q_i}) \cdot f_\theta(t_i) / \tau)}
{\sum_{j=1}^{N} \exp(f_\theta(q_i, o^{q_i}) \cdot f_\theta(t_j) / \tau)}
\end{split}
\label{eq:intra_loss}
\end{equation}
\begin{equation}
\begin{split}
\mathcal{L}_{\text{CM\_InfoNCE}} = \mathcal{L}_{disc} + \mathcal{L}_{gen} + \mathcal{L}_{intra}
\end{split}
\end{equation}
Here, $\mathcal{L}_{\text{disc}}$ and $\mathcal{L}_{\text{gen}}$ denote the discriminative and generative alignment losses respectively, and $\mathcal{L}_{\text{intra}}$ is the intra-mode alignment loss defined in Eq.~\ref{eq:intra_loss}. Therefore, these losses enforce mutual retrieval between generative and discriminative embeddings enabling flexible inference pathways. Meanwhile, the intra-mode alignment encourages generative embeddings to stay close to discriminative embeddings in the semantic space which prevents the semantic space of generative embeddings from drifting.

\subsection{Refine Reinforcement Learning}
% 这一段要强调disc和gen已经基本在一个空间内部，就可以把gen看成disc 的refine，鼓励产生更具有判别性的gen embedding，将其作为process reward，就可以省去一个reward model来批判rewrite process的对齐与否。

% 跨模态对齐机制已使生成式与判别式嵌入基本位于同一语义空间，我们可以将生成式嵌入视为判别式嵌入的精细化变体，直接以判别式嵌入为稳定参照，激励生成式重写产生更具判别性的表示。这一视角带来了关键优势：我们无需额外训练奖励模型来评判重写过程的对齐质量，而是将生成式嵌入是否比判别式嵌入具有更好的区分正负样本作为过程奖励信号，来监督rewrite过程是否带来了额外的信息增益。
% 同时我们保留了 之前工作设计的format reward来鼓励模型遵循预定好的rewrite的格式，保证过程的可解释性的同时保证生成式embedding的稳定提取，还保留了 基本的 gap reward，但是这个reward相对来说比较稀疏，奖励生成式embedding产生更大的相似度gap来啦开正负样本的距离。约定query为$q$，，reward function可以被写成如下的形式：
The cross-mode alignment mechanism maps both generative and discriminative embeddings into a unified semantic space, where generative embeddings serve as refined versions of discriminative ones. Using discriminative embeddings as stable anchors, we encourage the rewriting process to produce more informative representations. Specifically, we measure whether the generative embedding yields a larger similarity gap between positive and negative targets than the discriminative embedding  as a reward to supervise the information gain from rewriting.
Meanwhile, we retain the format reward to encourage adherence to a predefined rewriting template. We also keep the basic gap reward, which enlarges the similarity gap in generative embeddings to better separate positive and negative targets.
In summary, our overall reward function is defined as follows:
\begin{equation}
\mathcal{R}_{Format}(o) = \mathbb{I}\left( o \text{ adheres to Rewrite Prompt} \right)
\end{equation}
\begin{equation}
\mathcal{R}_{Gap}(o) = \mathbb{E}(f_\theta(q, o) \cdot f_\theta(t^+, {o^{t^+}_{i}})) - \mathbb{E}(f_\theta(q, o) \cdot f_\theta(t^-, o^{t^-}_{j}))
\end{equation}
\begin{equation}
\mathcal{R}_{Process}(o) = 
\begin{cases} 
1, & \text{if } \mathcal{R}_{Gap}(o) > f_\theta(q) \cdot f_\theta(t^+) - f_\theta(q) \cdot f_\theta(t^-), \\ 
0, & \text{if } \mathcal{R}_{Gap}(o) \leq f_\theta(q) \cdot f_\theta(t^+) - f_\theta(q) \cdot f_\theta(t^-), 
\end{cases}
\end{equation}
\begin{equation}
\mathcal{R}(o) = \mathcal{R}_{Format}(o) + \mathcal{R}_{Gap}(o) + \mathcal{R}_{Process}(o)
\end{equation}
where $q$ denote the query, $t^+$ and $t^-$ denote the positive and negative targets, respectively. $o$ represents the rewriting rollout generated by the model. $f_{\theta}(\cdot)$ denotes the normalized representation of $\langle \text{gen\_emb} \rangle$ or $\langle \text{disc\_emb} \rangle$, derived from the MLLM’s final-layer hidden state. $o^{t^+}_{i}$ and $o^{t^-}_{j}$ denote the rewriting rollouts for the corresponding positive and negative targets.
To optimize the rewriting policy, we adopt Group Relative Policy Optimization (GRPO). Specifically, for each query $q$, we sample a group of $K$ rewriting outputs $\{o_k\}_{k=1}^{K}$ from the old policy $\pi_{\theta_{\text{old}}}$. The policy $\pi_\theta$ is then optimized to maximize the following objective:
\begin{equation}
\begin{aligned}
\mathcal{L}_{\text{GRPO}}
&= \mathbb{E}_{q \sim \mathcal{D}, o_k \sim \pi_{\theta_{\text{old}}}}
\Bigg[
\frac{1}{K}\sum_{k=1}^{K}
\min \big( \frac{\pi_\theta(o_k|q)}
{\pi_{\theta_{\text{old}}}(o_k|q)}A_k, \\
&\qquad
\text{clip}(\frac{\pi_\theta(o_k|q)}
{\pi_{\theta_{\text{old}}}(o_k|q)},1-\epsilon,1+\epsilon)A_k \big)
 - \beta \mathbb{D}_{KL}(\pi_\theta \| \pi_{\text{ref}}) \Bigg]
\end{aligned}
\end{equation}
where $\pi_\theta$ and $\pi_{\theta_{\text{old}}}$ are the current and previous policies, $\pi_{\text{ref}}$ is a fixed reference, $\epsilon$ is the clipping coefficient, $\beta$ controls KL regularization, and $A_k$ is the group-normalized reward of the $k$-th rewrite within its group.

%% file: papers/4_Experiment.tex
\section{Experiments}

\input{tables/1_main}

\subsection{Implementation}

% Following VLM2Vec-V2, we use Qwen2-VL-2B-Instruct and Qwen2-VL-7B-Instruct as the backbone models. 同时，在 Rewrite SFT阶段，与UME-R1和VLM2Vec-V2基本可比的，我们使用了类似的数据，主要包括LLaVA-Hound，ViDoRe和VisRAG还有VLM2Vec提供的image相关数据，对应的具体数据配比，可以在附录详查。我们使用了同质数据组Batch，来引入一些较难的样本对比。
% $\lambda$被设置为1.0，the temperature $\tau$ is set to 0.02,the batch size to 512 (achieved through gradient accumulation)，训练一轮大约2500steps。学习率被设置为5e-5。在 Refine RL阶段，我们使用了大约15K的数据量，他们从三种模态中被均匀采样，同时尽可能保证没有在SFT过程中出现。Refine RL使用了GRPO的经典设置，具体的超参数包括group size $K = 8$, clipping parameter $\epsilon = 0.2$, and KL-divergence coefficient $\beta = 0.04$。训练一轮，学习率设置为5e-6。所有的实验都在32 $\times$ NVIDIA A100 (80GB) GPUs上完成。

Following VLM2Vec-V2~\cite{meng2025vlm2vec}, we employ Qwen2-VL-2B-Instruct and Qwen2-VL-7B-Instruct as the backbone models. Meanwhile, during the Rewrite SFT stage, we utilize approximately \textbf{1.5} Million training samples comparable to UME-R1 and VLM2Vec-V2, mainly including LLaVA-Hound, ViDoRe, VisRAG, and the image-based datasets in VLM2Vec. The detailed data composition  can be found in the Appendix. We adopt homogeneous batch groups to introduce comparisons with challenging samples.
The hyperparameter $\lambda$ is set to 1.0, the temperature $\tau$ is set to 0.02, and the batch size is configured as 512 (achieved through gradient accumulation). One training epoch consists of approximately 2500 steps, with the learning rate set to $5 \times 10^{-5}$. 
In the Refine RL stage, we use a dataset of around 15K samples, which are uniformly sampled from three modalities. Refine RL adopts the default GRPO settings with the following hyperparameters: group size $K = 8$, clipping parameter $\epsilon = 0.2$, and KL-divergence coefficient $\beta = 0.04$. For this stage, we train for one epoch with a learning rate of $5 \times 10^{-6}$.
All experiments are conducted on 32 $\times$ A800 GPUs.

\subsection{Datasets and Metrics}

% MMEB-V2作为最近兴起的评测模型多模态通用场景下检索能力的数据集，首先被我们选取来证明RIME的先进性和泛化性。如表\ref{tab:main_result}所示，MMEB-V2在Image和Video的检索场景下分别吧包含了36和18个子数据集，覆盖了基本的检索，分类，visual grounding，Moment Retrieval和视觉QA等等常见场景，采用Hit@1作为指标；在VisDoc 场景下，涉及复杂图表和论文图片等Document as Image的任务，采用NDCG@5作为指标。
% 同时为了测试RIME在更多更为专业和reasoning-intensive场景下的泛化性，我们选取了最新的MRMR和MR2-Bench来测试模型的复杂逻辑和专业知识情况下的检索表现，遵循各自的设置，我们采用nDCG@10作为指标，只针对于MRMR的Negation子数据集采用Hit@1作为指标。
% Additionally, 为了判断跨任务与跨领域视频检索场景下RIME的能力， we evaluate our models on UVRB, a universal video retrieval benchmark that systematically assesses model generalization across 16 subtasks.我们将采用平均精度均值 (mean Average Precision, mAP) 作为所有 UVRB 任务的主要评价指标。

In order to evaluate the retrieval capability of RIME, we adopt MMEB-V2, a recently proposed benchmark for evaluating multimodal universal retrieval capabilities. Unless otherwise specified, we use generative embeddings for evaluation.
As shown in Table~\ref{tab:main_result}, MMEB-V2 contains 36 and 18 sub-datasets for image and video  scenarios respectively, covering fundamental retrieval, classification, moment retrieval, visual QA, and other common tasks, with Hit@1 as the evaluation metric.
In the VisDoc scenario, which involves document-as-image tasks such as complex charts and paper figures, with NDCG@5 as the evaluation metric.

To further validate the generalization of RIME in more professional and reasoning-intensive scenarios, we adopt the recent MRMR specifically designed to assess the retrieval capability in expert-level, reasoning-intensive tasks.
Following their standard settings, we use nDCG@10 as the main metric. 

To assess the capability of RIME under cross-domain video retrieval, we evaluate our model on UVRB, a universal video retrieval benchmark that systematically assesses model generalization across 16 subtasks.
We mainly use Recall@1 as the primary evaluation metric, which aligns with the original setup of UVRB.

\subsection{Main Results}
\subsubsection{MMEB-V2 Benchmark}
% 表\ref{tab:main_result}展示了在MMEB-V2基准上的对比结果，涵盖图像、视频和视觉文档三种模态的78项任务。RIME在开源模型中取得了最先进的成果：RIME-7B以68.3分的综合成绩，分别超越UME-R1-7B 3.8分、VLM2Vec-7B 16.0分。从不同模态来看，RIME-7B在图像任务上获得73.4分（较UME-R1提升2.1分），视频任务49.4分（提升1.9分），视觉文档任务75.6分（提升8.5分）。值得注意的是，基于Qwen2-VL的RIME综合成绩甚至和Think-Then-Embed相当，但Think-Then-Embed却采用了额外的Reasoner来推导CoT。以上皆可证明，the rewrite-driven paradigm 不仅做到了相比于Cot-based model更好的性能，而且更好的兼容了自回归语言建模和embedding对比学习，一定程度上打破了generative embedding的生成与嵌入目标的联合优化存在梯度冲突限制。

% 与缺乏重写机制的DUME相比，RIME-7B提升了12.4分（68.3分对比55.9分），验证了检索友好型重写机制带来的信息增益。与联合优化生成和嵌入但是最终还是选取静态embedding提取的CAFe相比，RIME-7B提升了12.4分（68.3分对比55.9分）
Table \ref{tab:main_result} presents the comparative results on the MMEB-V2 benchmark, a comprehensive evaluation suite for multimodal embedding models. Under comparable settings, RIME achieves state-of-the-art results among open-source models: RIME-7B (Qwen2-VL) attains a comprehensive score of 68.6, surpassing UME-R1-7B by 4.1 points and VLM2Vec-7B by 16.3 points.
Across distinct modalities, RIME-7B demonstrates consistent advantages, achieving 73.4 on image-based tasks (+2.1 over UME-R1), 49.4 on video-based tasks (+1.9), and 75.6 on visual document tasks (+8.5). Notably, RIME-7B matches the performance of Think-Then-Embed at 68.6 without requiring an additional Reasoner module, while also outperforming CAFe by 8.0 points—a model that jointly optimizes generation and embedding yet still extracts static embeddings.
These results collectively indicate that RIME not only surpasses Chain-of-Thought-based models, but also achieves a more effective reconciliation between autoregressive language modeling and embedding contrastive learning.

% Compared to DUME, which lacks a rewrite mechanism, RIME-7B improves by 12.7 points (68.6 vs. 55.9), validating the information gain from retrieval-friendly rewriting. Compared to CAFe, which jointly optimizes generation and embedding but still extracts static embeddings, RIME-7B improves by 8.0 points (68.6 vs. 60.6), demonstrating the potential for deeper

\input{tables/2_MRMR}
\input{tables/5_ablation_component}

\subsubsection{MRMR Benchmark}

Table~\ref{tab:mrmr} presents the comparative results on the MRMR benchmark, covering 11 expert-level reasoning-intensive subtasks across three categories: Knowledge, Theorem, and Contradiction. This benchmark sufficiently demonstrates the generalizability of RIME in reasoning-intensive retrieval tasks.
RIME-7B achieves the best comprehensive score of 50.2 among all evaluated models, surpassing UME-R1-7B by 2.2 points, Ops-MM-Embed-7B by 2.1 points, and GME-7B by 14.0 points.
In Knowledge-intensive tasks, RIME-7B excels in Medicine (58.5, +2.8 over UME-R1), Science (73.6, +0.7), and Humanities (71.4, +7.3), while maintaining competitive performance in Art.
In Theorem reasoning tasks, RIME-7B consistently outperforms all baselines, achieving 29.2 on Math (+2.0), 43.0 on Physics (+3.8), 35.6 on Engineering (+3.4), and 51.8 on Business (+4.0).
These results validate that the rewrite-driven paradigm effectively enhances reasoning capability in complex expert-level retrieval tasks by activating internal knowledge, rather than relying on CoT reasoning to derive answers. This is further supported by the ablation study in Table~\ref{tab:ablation_rewrite_vs_cot}.

\subsubsection{UVRB Benchmark}

% 表~\ref{tab:uvrb} 展示了在 UVRB 基准测试上的对比结果。UVRB 是一个通用视频检索基准，涵盖 16 个子任务，横跨不同任务类型和复杂场景，RIME在可比训练数据量的情况下大幅超越了之前的模型，验证了RIME在视频场景下的通用检索适配能力。
% 具体而言，在可比设置下，RIME-7B 在开源模型中取得了最佳综合性能 55.6，超越 Unite-7B 1.8 分，超越 GME-7B 2.6 分。
% 值得注意的是，RIME 在长上下文场景中展现出显著优势，超越了所有基线模型，甚至优于使用额外数据的 GVE。
% 这验证了结构化重写范式能够有效捕捉视频内容中的时序动态和长程依赖关系。
% 在粗粒度检索任务中，RIME-7B 取得 56.2 分，较 Unite-7B 提升 2.1 分，表明重写机制为通用视频理解提供了更好的语义抽象能力。
% 在时序子领域任务中，RIME-7B 达到 42.7 分，为所有基线模型中最高，证明了模态感知描述性重构在捕捉视频时序动态方面的有效性。

Table~\ref{tab:uvrb} presents the comparative results on the UVRB benchmark, a universal video retrieval benchmark that systematically evaluates model generalization across 16 subtasks spanning diverse task types and application domains. 
Under comparable training data settings, RIME significantly outperforms prior state-of-the-art models, validating its strong general retrieval adaptation capability in video-centric scenarios.
Specifically, RIME-7B achieves the best overall performance with a comprehensive score of 55.6, surpassing Unite-7B by 1.8 points and GME-7B by 2.6 points among all evaluated models. 
Notably, RIME demonstrates particularly strong performance in long-context video understanding scenarios, where it exceeds all baseline models and even outperforms GVE, which leverages additional training data. 
On coarse-grained retrieval tasks that require high-level semantic abstraction, RIME-7B achieves a score of 56.2, representing a 2.1-point improvement over Unite-7B. 
Furthermore, on temporal-sensitive subtasks that demand fine-grained temporal reasoning, RIME-7B reaches 42.7 points, the highest among all baselines.
Overall, these results demonstrate that the structured rewriting paradigm can effectively enable more robust and generalizable video retrieval across diverse scenarios and task complexities.

\subsection{Analysis}
This section presents ablation studies to validate the design of RIME. Unless otherwise specified, experiments use Qwen2-VL-2B-Instruct with generative embedding mode. We evaluate each core component (subsection 4.4.1), compare retrieval-friendly rewriting with chain-of-thought reasoning (subsection 4.4.2), and analyze mutual retrievability enabled by CMA (subsection 4.4.3).

% \input{tables/4_MR2}
\input{tables/9_ablation_rewrite_vs_cot}
\input{tables/3_URVB}

\subsubsection{Ablation on Core Components}

% 为验证各提出训练pipeline中各个贡献点的有效性，我们通过向基线模型逐步添加重写驱动的联合监督微调、跨模态对齐和精炼强化学习进行了消融实验。如表~\ref{tab:ablation_component}所示，每个组件均对整体性能提升有所贡献。仅重写机制就为基线模型带来了显著增益，证明了检索友好型重写的优越性和联合训练生成和嵌入相对于纯判别式嵌入的信息增益。
% 跨模态对齐通过对齐判别式与生成式嵌入空间实现了两种模式的互相可检索，带来了test time的便利性的同时，也获得了一定的性能增益，说明生成式嵌入能从判别式嵌入中寻求对齐来减少重写噪声。
% 最后，精炼强化学习通过优化重写过程以生成更合理的重写过程来鼓励对于生成思考过程的进一步细化，在MMEB-V2上获得了一定的优化提升。

To validate the contribution of each component in the proposed training pipeline, we conduct ablation studies by progressively integrating Rewrite-Driven Joint SFT, Cross-Mode Alignment (CMA), and Refine-RL into the baseline framework. As shown in Table~\ref{tab:ablation_component}, each component yields measurable improvements to the overall performance.
The Rewrite mechanism alone brings substantial gains of 1.9, demonstrating the effectiveness of retrieval-friendly rewriting and the advantages of jointly training generation and embedding tasks over relying solely on discriminative embeddings. CMA aligns discriminative and generative embedding spaces, enabling mutual retrievability between the two modalities. This enhancement improves test-time convenience while contributing a modest improvement to overall performance. Refine-RL further refines the rewriting process by optimizing the model to generate more coherent and effective rewriting procedures, achieving the best overall performance of 64.1 on the MMEB-V2 benchmark.

\subsubsection{Ablation on Rewrite vs. Chain-of-Thought}

% 为验证重写范式相较于链式推理（CoT）的有效性，我们在MMEB-V2上开展了对比不同思考过程的自消融研究。如表\ref{tab:ablation_rewrite_vs_cot}所示，在非强化学习变体中，重写配置取得了最佳性能（综合指标：63.5），显著优于CoT（59.1），高出4.4个百分点。值得注意的是，移除强制总结答案对两种范式均有提升：CoT（无总结答案）达到60.5（较CoT提升1.4），而重写（无总结答案）则比重写（有总结答案）高出2.3个百分点（63.5 vs. 61.2）。这证实了在多样化检索任务中，答案级别的总结会带来粒度不匹配问题，因为检索查询通常缺乏唯一对应的答案。

% 从效率角度看，重写仅使用平均212个token——不到CoT（475个token）的一半——同时实现了更优性能，这表明面向检索友好的重写相比逐步推理能提供更紧凑且语义保真的表示。Refine-RL进一步将重写性能提升至64.1（+0.6），同时token数仅适度增加至232，验证了奖励引导的优化能有效精炼重写过程，从而生成更具信息量的生成式嵌入。

To validate the effectiveness of rewriting relative to chain-of-thought reasoning, we conduct comprehensive ablation studies on the MMEB-V2 benchmark. As shown in Table~\ref{tab:ablation_rewrite_vs_cot}, Rewrite achieves the best performance among non-RL variants, substantially outperforming CoT by 4.4 points. Removing the forced summarized answer improves both paradigms: CoT (w/o ans.) attains 60.5, a gain of 1.4 points, while Rewrite surpasses Rewrite (w/ ans.) by 2.3 points. This confirms that summarized answers introduce extra retrieval noise, and mismatched answers can mislead generative embeddings.
From an efficiency perspective, Rewrite uses only 212 average tokens while achieving superior performance, demonstrating that retrieval-friendly rewriting provides a more compact yet semantically faithful representation than step-by-step reasoning. After applying Refine-RL, we achieve an average token reduction of approximately 52\% compared to the UME-R1 baseline.

% \begin{figure}[!t]
%     \centering
%     \includegraphics[width=\columnwidth]{figures/compare_exp.pdf}
%     \caption{
%     An intuitive comparison of the rewrite process versus CoT in terms of retrieval similarity.
%     }
%     \label{fig:compare}
% \end{figure}

\subsubsection{Ablation on Two Embedding Modes}

% ==================== 中文版本 ====================
% 基于 Qwen2-VL-7B 基模，我们在推理阶段对四种嵌入模式组合进行了消融实验，以验证跨模态对齐（CMA）所实现的判别式与生成式嵌入之间的互检索能力。
% 如表~\ref{tab:ablation_mode}所示，Gen.-Gen. 配置在三个基准上均取得最佳性能（MMEB-V2: 68.6, MRMR: 50.2, UVRB: 55.6, Avg.: 58.1），验证了重写机制带来的语义增强效果。
% 值得注意的是，在引入 CMA 之后，即使不使用生成式嵌入（第二行 Disc.-Disc.），综合性能也从纯判别式对比的 55.8 提升至 56.3，说明 CMA 的训练目标本身也对判别式嵌入产生了正则化效果。
% 混合模式同样表现优异：Disc.-Gen. 达到 57.0（0 查询 token），Gen.-Disc. 达到 57.1（232 查询 token）。
% 从实际部署角度，Disc.-Gen. 模式尤为实用——目标库可离线预编码为生成式嵌入以增强语义表达，查询端则直接使用判别式嵌入实现零额外推理开销，同时相比纯判别式模式仍有 1.2 分的提升。

% ==================== English Version ====================

To validate the mutual retrievability enabled by CMA, we conduct ablation studies on four embedding mode combinations during inference using the Qwen2-VL-7B backbone. As shown in Table~\ref{tab:ablation_mode}, the Gen.-Gen. configuration achieves the best performance across all benchmarks, with an average score of 58.1, validating the semantic enhancement brought by the rewriting mechanism.
Notably, after introducing CMA, the pure discriminative mode improves from 55.8 to 56.3 on average, suggesting that CMA serves as an effective regularizer for discriminative embeddings. Hybrid modes also demonstrate competitive performance: Disc.-Gen. reaches 57.0 with zero query tokens, while Gen.-Disc. achieves 57.1 with 232 query tokens.
From a deployment perspective, Disc.-Gen. presents a particularly attractive and flexible solution: the corpus can be offline encoded using generative embeddings to capture richer semantics, while queries leverage discriminative embeddings at zero inference cost, enabling efficient and scalable retrieval in practice.
% , still yielding a 1.2-point improvement over the pure discriminative mode.   

%% file: tables/1_main.tex
\begin{table*}[t]
\caption{Results on the MMEB-V2 benchmark. The results in \textbf{bold} represent the best performance among different model sizes. CLS: classification, QA: question answer, RET: retrieval, GD: grounding, MRET: moment retrieval, VDR: ViDoRe, VR: VisRAG, OOD: out-of-distribution. Reported metrics adhere to the settings of VLM2Vec-V2. Detailed results can be found in Appendix.}

\centering
\renewcommand{\arraystretch}{1.05}
\resizebox{\textwidth}{!}{
\begin{tabular}{ll ccccc ccccc ccccc c}
\toprule
\multirow{2}{*}{\textbf{Model}} 
& \multirow{2}{*}{\textbf{Backbone}}
& \multicolumn{5}{c}{\textbf{Image}} 
& \multicolumn{5}{c}{\textbf{Video}} 
& \multicolumn{5}{c}{\textbf{VisDoc}}
& \multirow{2}{*}{\textbf{All}}\\
\cmidrule(lr){3-7} \cmidrule(lr){8-12} \cmidrule(lr){13-17}
& & \textbf{CLS} & \textbf{QA} & \textbf{RET} & \textbf{GD} & \textbf{Avg.} 
& \textbf{CLS} & \textbf{QA} & \textbf{RET} & \textbf{MRET} & \textbf{Avg.} 
& \textbf{VDRv1} & \textbf{VDRv2} & \textbf{VR} & \textbf{OOD} & \textbf{Avg.} & \\
\midrule
\textbf{\# of Datasets} &
& 10 & 10 & 12 & 4 & 36 
& 5 & 5 & 5 & 3 & 18 
& 10 & 4 & 6 & 4 & 24
& 78 
\\
\midrule
\multicolumn{18}{c}{\emph{Closed-Source or w/ Additional Data}} \\
\midrule
Think-Then-Embed & Qwen2-VL & 
69.7 & 72.4 & 74.0 & 90.6 & 74.2 & 
49.1 & 60.6 & 36.4 & 37.2 & 46.8 & 
84.1 & 62.7 & 91.9 & 47.6 & 76.4 & 
68.6 \\
Ops-MM-Embed & Qwen2-VL & 
69.7 & 69.6 & 73.1 & 87.2 & 72.7 & 
59.7 & 62.2 & 45.7 & 43.2 & 53.8 & 
80.1 & 59.6 & 79.3 & 67.8 & 74.4 & 
68.9 \\
Seed-1.6-embedding & Seed1.6-flash & 
 76.1 & 74.0 & 77.9 & 91.3 & 77.8 & 55.0 & 60.9 & 51.3 & 53.5 & 55.3 & 85.5 & 56.6 & 84.7 & 43.1 & 73.4 & 71.3 \\
Qwen3-VL-Embedding & Qwen3-VL & 
74.2 & 81.1 & 80.2 & 92.3 & 80.1 & 
78.4 & 71.0 & 58.7 & 56.1 & 67.1 & 
87.2 & 69.9 & 88.7 & 73.3 & 82.4 & 
77.8 \\
\midrule
\multicolumn{18}{c}{\emph{$\sim$ 2B Model Size}} \\
\midrule
ColPali-V1.3 & PaliGemma & 
40.3 & 11.5 & 48.1 & 40.3 & 34.9 & 
26.7 & 37.8 & 21.6 & 25.5 & 28.2 & 
83.6 & 52.0 & 81.1 & 43.1 & 71.0 & 
44.4 \\
GME & Qwen2-VL & 
54.4 & 29.9 & 66.9 & 55.5 & 51.9 & 
34.9 & 42.0 & 25.6 & 32.4 & 33.9 &
\textbf{86.1} & \textbf{54.0} & \textbf{82.5} & 43.1 & \textbf{72.7} &
54.1 \\
VLM2Vec & Qwen2-VL & 
58.7 & 49.3 & 65.0 & 72.9 & 59.7 & 
33.4 & 30.5 & 20.6 & 33.0 & 29.0 & 
49.8 & 13.5 & 51.8 & 33.5 & 41.6 & 
47.0 \\
VLM2Vec-V2 & Qwen2-VL & 
62.9 & 56.3 & 69.5 & 77.3 & 64.9 &
39.3 & 34.3 & 28.8 & 38.5 & 34.9 & 
75.5 & 44.9 & 79.4 & 39.4 & 65.4 & 
58.0 \\
DUME & Qwen2-VL & 
59.3 & 55.0 & 66.3 & 78.0 & 62.5 &
37.7 & 46.6 & 17.1 & 30.0 & 33.2 & 
67.6 & 43.3 & 47.1 & 33.8 & 52.8 & 
52.7 \\
UME-R1 & Qwen2-VL & 
64.8 & 62.8 & 67.6 & 77.2 & 66.6 &
44.3 & 51.2 & 32.9 & \textbf{39.7} & 42.2 & 
72.4 & 46.2 & 79.2 & 37.2 & 63.9 & 
60.1 \\
\rowcolor[HTML]{E7E7F9}
\textbf{RIME (Ours)} & Qwen2-VL & 
\textbf{67.9} & \textbf{64.4} & \textbf{69.8} & \textbf{82.1} & \textbf{69.1} & 
\textbf{48.0} & 
\textbf{52.1} &
\textbf{33.6} &
39.2 &
\textbf{43.7} &
76.4 & 51.4 & 81.7 & \textbf{63.9} & 71.4 & 
\textbf{64.1} \\
\midrule
\multicolumn{18}{c}{\emph{$\sim$ 7B Model Size}} \\
\midrule
GME & Qwen2-VL & 
57.7 & 34.7 & 71.2 & 59.3 & 56.0 & 
37.4 & 50.4 & 28.4 & 38.2 & 38.6 & 
\textbf{89.4} & \textbf{55.6} & 85.0 & 44.4 & 75.2 & 
57.8 \\
LamRA & Qwen2.5-VL & 
51.7 & 34.1 & 66.9 & 56.7 & 52.4 & 
32.9 & 42.6 & 23.2 & 37.6 & 33.7 & 
56.3 & 33.3 & 58.2 & 40.1 & 50.2 & 
47.4 \\
VLM2Vec & Qwen2-VL & 
62.7 & 56.9 & 69.4 & 82.2 & 65.5 & 
39.1 & 30.0 & 29.0 & 40.6 & 34.0 & 
56.9 & 9.4 & 59.1 & 38.1 & 46.4 & 
52.3 \\
DUME & Qwen2-VL & 
64.2 & 57.0 & 70.8 & 81.8 & 66.4 &
32.9 & 47.4 & 8.6 & 28.0 & 29.4 & 
67.1 & 35.2 & 82.6 & 34.9 & 60.3 & 
55.9 \\
CAFe & LLaVA-OV & 
63.6 & 61.7 & 69.1 & \textbf{87.6} & 67.6 & 
35.8 & 58.7 & 34.4 & 39.5 & 42.4 & 
70.7 & 49.6 & 79.5 & 38.1 & 63.9 & 
60.6 \\
UME-R1 & Qwen2-VL & 
67.1 & 69.2 & 71.9 & 84.9 & 71.3 &
48.6 & 60.7 & 38.2 & 39.3 & 47.5 & 
75.7 & 50.5 & 83.7 & 37.6 & 67.1 & 
64.5 \\
% \rowcolor[HTML]{E7E7F9}
% \textbf{RIME (Ours)} & Qwen3-VL & 
% 66.1 & \textbf{72.3} & 70.8 & \textbf{91.4} & 72.0 &
% 50.5 & \textbf{63.4} & \textbf{39.6} & \textbf{44.8} & \textbf{50.0} & 
% 80.7 & 49.8 & 85.1 & 64.9 & 74.0 & 
% 67.5 \\
\rowcolor[HTML]{E7E7F9}
\textbf{RIME (Ours)} & Qwen2-VL & 
\textbf{70.3} & \textbf{71.7} & \textbf{73.2} & 86.3 & \textbf{73.4} &
\textbf{52.6} & \textbf{62.0} & \textbf{38.4} & \textbf{41.6} & \textbf{49.4} &
\textbf{80.9} & \textbf{55.6} & \textbf{85.8} & \textbf{66.9} & \textbf{75.6} & 
\textbf{68.6} \\

\bottomrule
\end{tabular}
}
\label{tab:main_result}
\end{table*}

%% file: tables/2_MRMR.tex
% UME-R1 复现标记给一下！！！
\begin{table*}[t]
\caption{Results on the \textbf{MRMR} benchmark. We report nDCG@10 for all subtasks except Negation, for which we use Hit@1: Art, Medicine (Med.), Science (Sci.), Humanities (Hum.), Math, Physics (Phy.), Engineering (Eng.), Business (Bus.), Negation (Neg.), Design (Des.), and Traffic (Tra.). All denotes the average score across 11 subtasks.}
\label{tab:mrmr}
\centering
\scriptsize
\setlength{\tabcolsep}{4.5pt}
\renewcommand{\arraystretch}{0.98}

\resizebox{\textwidth}{!}{%
\begin{tabular}{llc cccc cccc ccc c}
\toprule
\multirow{2}{*}{\textbf{Model}} 
& \multirow{2}{*}{\textbf{Backbone}}
& \multirow{2}{*}{\textbf{Size}}
& \multicolumn{4}{c}{\textbf{Knowledge}} 
& \multicolumn{4}{c}{\textbf{Theorem}} 
& \multicolumn{3}{c}{\textbf{Contradiction}} 
& \multirow{2}{*}{\textbf{All}} \\
\cmidrule(lr){4-7} \cmidrule(lr){8-11} \cmidrule(lr){12-14}
& & & Art & Med. & Sci. & Hum. & Math & Phy. & Eng. & Bus. & Neg. & Des. & Tra. & \\
\midrule
\multicolumn{15}{c}{\emph{Baseline Models}} \\
\midrule
EVA-CLIP & EVA-ViT & 0.4B &
10.2 & 13.5 & 26.1 & 12.9 & 6.2 & 10.5 & 9.3 & 11.7 & 8.5 & 4.4 & 5.4 & 10.8 \\
OpenCLIP & ViT-G/14 & 1B &
56.0 & 17.9 & 33.2 & 22.0 & 5.7 & 5.0 & 7.0 & 9.7 & 13.0 & 8.1 & 12.4 & 17.3 \\
VISTA & Qwen2-VL & 2B &
21.3 & 27.8 & 32.6 & 17.0 & 18.8 & 17.1 & 17.3 & 28.6 & \textbf{20.0} & 20.2 & 9.4 & 20.9 \\
E5-V & LLaVA-Next & 8B &
25.1 & 11.7 & 16.6 & 10.8 & 2.1 & 3.4 & 2.5 & 5.2 & 11.5 & 3.7 & 2.1 & 8.6 \\
VLM2Vec & Qwen2-VL & 7B &
53.5 & 22.4 & 36.7 & 24.0 & 2.1 & 2.8 & 2.8 & 2.9 & 11.5 & 5.6 & 18.3 & 18.1 \\
ColPali & PaliGemma & 3B & 36.1 & 29.9 & 42.7 & 29.2 & 5.7 & 14.8 & 12.0 & 24.6 & 28.5 & 19.4 & 18.2 & 23.7 \\
GME & Qwen2-VL & 7B &
54.3 & 40.1 & 46.8 & 45.6 & 28.8 & 36.0 & 30.2 & 45.1 & 15.0 & 26.3 & 29.6 & 36.2 \\
MM-Embed & NV-Embed & 8B &
65.6 & 53.0 & 63.5 & 62.8 & 23.6 & 30.8 & 27.4 & 44.9 & 7.0 & 23.8 & 34.9 & 39.8 \\
Ops-MM-Embed & Qwen2-VL & 7B &
\textbf{79.3} & 52.5 & 70.0 & 67.8 & 27.7 & 39.5 & 30.1 & 52.3 & 8.0 & 55.9 & \textbf{45.8} & 48.1 \\
UME-R1 & Qwen2-VL & 7B &
77.8 & 55.7 & 72.9 & 64.1 & 27.2 & 39.2 & 32.2 & 47.8 & 7.5 & 61.9 & 41.7 & 48.0 \\
\rowcolor[HTML]{E7E7F9}
\textbf{RIME (Ours)} & Qwen2-VL & 7B &
76.8 & \textbf{58.5} & \textbf{73.6} & \textbf{71.4} & \textbf{29.2} & \textbf{43.0} & \textbf{35.6} & \textbf{51.8} & 8.5 & \textbf{64.1} & 39.5 & \textbf{50.2} \\
% \rowcolor[HTML]{E7E7F9}
% \textbf{RIME (Ours)} & Qwen3-VL & 8B &
% 75.4 & \textbf{62.6} & 73.5 & \textbf{73.8} & \textbf{36.7} & \textbf{49.6} & \textbf{40.8} & \textbf{56.2} & 8.0 & 55.1 & 37.4 & \textbf{51.7} \\
\bottomrule
\end{tabular}%
}
\end{table*}

%% file: tables/5_ablation_component.tex
% ----------------- Ablation on Core Components -----------------
% RL 给一个MRMR的结果
\begin{table}[t]
\caption{Ablation study on core components of RIME. We report the average scores on MMEB-V2 benchmark. All crosses (\ding{55}\ding{55}\ding{55}) denote pure discriminative embedding training.}

\centering
\small
\setlength{\tabcolsep}{4.5pt}

\resizebox{\columnwidth}{!}{%
\begin{tabular}{ccc cccc}
\toprule
\textbf{Rewrite SFT} & \textbf{CMA} & \textbf{Refine-RL} & \textbf{Image} & \textbf{Video} & \textbf{VisDoc} & \textbf{All} \\
\midrule
\ding{55} & \ding{55} & \ding{55} & 65.5 & 41.1 & 68.0 & 61.3 \\
\ding{51} & \ding{55} & \ding{55} & 68.2 & 43.2 & 70.0 & 63.2 \\
\ding{51} & \ding{51} & \ding{55} & 68.6 & 43.1 & 70.6 & 63.5 \\
\rowcolor[HTML]{E7E7F9}
\ding{51} & \ding{51} & \ding{51} & \textbf{69.1} & \textbf{43.7} & \textbf{71.4} & \textbf{64.1} \\
\bottomrule
\end{tabular}
}
\label{tab:ablation_component}
\end{table}

%% file: tables/9_ablation_rewrite_vs_cot.tex
% ----------------- Ablation: Rewrite vs. CoT -----------------
\begin{table}[t]
\caption{Comparison between the rewrite and CoT reasoning on MMEB-V2 benchmark. All CoT variants are trained and evaluated based on UME-R1 for fair comparison.}

\centering

\resizebox{\columnwidth}{!}{%
\begin{tabular}{l ccccc}
\toprule
\textbf{Thinking Process} & \textbf{Image} & \textbf{Video} & \textbf{VisDoc} & \textbf{All} & \textbf{Avg. Tokens}\\
\midrule
CoT &  65.2 & 41.2 & 63.5 & 59.1 & 475 \\
CoT (w/o ans.) & 65.6 & 41.0 & 66.0 & 60.5 & 417 \\
Rewrite (w/ ans.) & 64.3 & 41.3 & 68.2 & 61.2 & 246 \\
Rewrite & 68.6 & 43.1 & 70.6 & 63.5 & \textbf{212} \\
\rowcolor[HTML]{E7E7F9}
Rewrite (w/ Refine-RL) & \textbf{69.1} & \textbf{43.7} & \textbf{71.4} & \textbf{64.1} & 232 \\
\bottomrule
\end{tabular}
}

\label{tab:ablation_rewrite_vs_cot}
\end{table}

%% file: tables/3_URVB.tex
\begin{table*}[t]
\caption{Results on the \textbf{UVRB} benchmark. We mainly report Recall@1 for three \textbf{Tasks}: Textual (Txt.), Composed (Cmp.), and Visual (Vis.); three \textbf{Domains}: Coarse-Grained (Coarse-G.), Fine-Grained (Fine-G.), and Long-Context (Long-Ctx.); and three \textbf{Sub-domains}: Spatial (Spa.), Temporal (Temp.), and Partially Relevant (PR.). \textbf{All} is the average across all 16 datasets.}

\centering
\tiny 
\setlength{\tabcolsep}{4.5pt}
\renewcommand{\arraystretch}{0.9}
\resizebox{\textwidth}{!}{%
\begin{tabular}{llc ccc ccc ccc c}
\toprule
\multirow{2}{*}{\textbf{Model}} 
& \multirow{2}{*}{\textbf{Backbone}}
& \multirow{2}{*}{\textbf{Size}}
& \multicolumn{3}{c}{\textbf{Tasks}} 
& \multicolumn{3}{c}{\textbf{Domains}} 
& \multicolumn{3}{c}{\textbf{Sub-domains}} 
& \multirow{2}{*}{\textbf{All}} \\
\cmidrule(lr){4-6} \cmidrule(lr){7-9} \cmidrule(lr){10-12}
& & & Txt.  & Cmp. & Vis. & Coarse-G. & Fine-G.  & Long-Ctx. & Spa. & Temp. & PR. & \\
\midrule
\multicolumn{13}{c}{\emph{Closed-Source or w/ Additional Data}} \\
\midrule

GVE & Qwen2-VL & 7B &
65.7 & 31.2 & 65.7 & 58.7 & 57.0 & 81.4 & 82.1 & 46.9 & 41.9 & 57.3 \\
\midrule
\multicolumn{13}{c}{\emph{Baseline Models}} \\
\midrule
ViCLIP & ViT-L/14 & 0.4B & 33.6 & 26.3 & 64.0 & 38.0 & 31.5 & 31.3 & 48.4 & 28.9 & 17.1 & 35.2 \\
BGE-VL & Qwen2-VL & 2B &
49.7 & 26.8 & 62.2 & 44.8 & 40.6 & 63.6 & 66.4 & 29.2 & 26.1 & 44.3 \\
Unite & Qwen2-VL & 2B &
53.6 & 24.2 & 65.4 & 45.5 & 47.1 & 68.1 & 72.5 & 34.7 & 34.1 & 48.0 \\
VLM2Vec-V2 & Qwen2-VL & 2B &
58.7 & 26.3 & 61.3 & 49.8 & 50.2 & 76.2 & 80.9 & 34.8 & 34.8 & 50.8 \\
B3 & Qwen2-VL & 7B &
57.0 & 27.0 & 67.8 & 48.2 & 50.5 & 72.2 & 79.7 & 36.4 & 35.5 & 51.1 \\
UniME & Qwen2-VL & 7B &
56.1 & 30.8 & \textbf{70.2} & 50.0 & 51.8 & 66.4 & 78.5 & 39.6 & 37.3 & 52.1 \\
GME & Qwen2-VL & 7B &
60.4 & \textbf{34.1} & 61.5 & 51.8 & 50.7 & 78.8 & 74.9 & 37.3 & 39.8 & 53.0 \\
Unite & Qwen2-VL & 7B &
\textbf{60.9} & 25.4 & 66.6 & 54.1 & 53.9 & 74.6 & 77.9 & 41.2 & \textbf{42.5} & 53.8 \\
UME-R1 & Qwen2-VL & 7B & 59.1 & 15.9 & 66.4 & 55.5 & 53.9 & \textbf{82.5} & \textbf{82.5} & 42.7 & 41.8 & 54.6 \\
\rowcolor[HTML]{E7E7F9}
\textbf{RIME (Ours)} & Qwen2-VL & 7B &
60.8 & 22.0 & 67.5 & \textbf{56.2} & \textbf{54.0} & 81.6 & 79.9 & \textbf{43.6} & 41.3 & \textbf{55.6} \\

\bottomrule
\end{tabular}
}
\label{tab:uvrb}
\end{table*}

%% file: papers/5_Conclusion.tex
\section{Conclusion}
\input{tables/7_ablation_embedding_mode}

% ==================== 中文版本 ====================
% 本文提出 RIME，通过检索友好的重写范式替代思维链推理，实现了高效且泛化性强的生成式多模态嵌入。实验表明 RIME 在MMEB-V2,MRMR,UVRB基准上取得最先进性能，同时大幅降低推理开销。
% 然而，当前框架仍存在局限性, Rewrite虽然相对来说优化了原有冗余的思考过程和不必要的总结式回答，但他依旧是带来了很大的推理延迟，这是生成式嵌入不可能避免的弊病，对于实际规模检索场景是不可接受的。但是Latent思考过程或许是解决方案之一，后续的工作或许可以从此入手进一步的把生成式embedding做的实时性更强，更为通用的适配到所有场景。

% ==================== English Version ====================
In this paper, we propose RIME, which directly substitutes traditional chain-of-thought reasoning with a novel retrieval-friendly rewriting paradigm, yielding efficient yet highly generalizable generative multimodal embeddings. Extensive experiments demonstrate that RIME achieves state-of-the-art performance across the MMEB-V2, MRMR, and UVRB benchmarks, while simultaneously reducing inference overhead significantly.
Nevertheless, the proposed framework still exhibits certain limitations. Although the Rewrite mechanism alleviates the original redundant reasoning process and unnecessary summarized answers to a certain degree, it nonetheless introduces considerable inference latency. This represents an inherent limitation of generative embedding models, which is difficult to tolerate in practical large-scale retrieval applications. Accordingly, latent thinking processes may provide a promising direction for alleviating this issue.
We aim to further convert explicit reasoning steps into an implicit latent form, facilitating more practical deployment in real-world retrieval scenarios.

%% file: tables/7_ablation_embedding_mode.tex
% ----------------- Ablation on Embedding Mode -----------------
\begin{table}[t]
\caption{Ablation study on embedding modes during inference. Disc. denotes discriminative embeddings, Gen. denotes generative embeddings. Qry. Tokens refers to the average tokens generated per query at retrieval time.}
\label{tab:ablation_mode}
\centering

\renewcommand{\arraystretch}{1.0}
\resizebox{\columnwidth}{!}{%
\begin{tabular}{ll cccc c}
\toprule
\textbf{Query} & \textbf{Target} & \textbf{MMEB-V2} & \textbf{MRMR} & \textbf{UVRB} & \textbf{Avg.} & \textbf{Qry. Tokens} \\
\midrule
\multicolumn{7}{c}{\emph{Pure Discriminative Contrast}} \\
\midrule
Disc. & Disc. & 66.9 & 48.0 &  52.6  & 55.8 & \textbf{0} \\
\midrule
\multicolumn{7}{c}{\emph{Cross Alignment with two modes}} \\
\midrule
Disc. & Disc. & 67.4 & 48.4 & 53.0 & 56.3 & \textbf{0} \\

Disc. & Gen. & 67.9 & 49.9 & 52.8 & 57.0 & \textbf{0} \\

Gen. & Disc. & 68.0 & 49.1 & 54.2 & 57.1 & 232 \\
\rowcolor[HTML]{E7E7F9}
Gen. & Gen. & \textbf{68.6} & \textbf{50.2} & \textbf{55.6} & \textbf{58.1} & 232 \\
\bottomrule
\end{tabular}
}
\end{table}

%% file: appendix/appendix.tex
% \documentclass[sigconf, screen, review, anonymous]{acmart}

% \sloppy

% \usepackage{fontawesome}

% \usepackage{multirow} 

% \usepackage{bbm}

% \usepackage{algorithm}
% \usepackage{algorithmic}
% \usepackage{amsmath}
% \usepackage{xcolor}
% \usepackage{booktabs}
% \usepackage{graphicx}
% \let\Bbbk\relax
% \usepackage{amssymb}
% \usepackage{colortbl}
% \usepackage{array}
% \usepackage{makecell}
% \usepackage{float}
% \floatplacement{table}{H} 
% \usepackage{subcaption}
% \usepackage{pifont}
% \usepackage{longtable}
% \usepackage{balance}

% \AtBeginDocument{%
%   \providecommand\BibTeX{{%
%     Bib\TeX}}}

% \setcopyright{acmlicensed}
% \copyrightyear{2025}
% \acmYear{2025}
% \acmDOI{XXXXXXX.XXXXXXX}

% \acmConference[MM'26]{ACM Multimedia}{November 2026}{Rio de Janeiro, Brazil}

% \acmISBN{978-1-4503-XXXX-X/2018/06}

% \acmSubmissionID{965}

% \settopmatter{printacmref=false}     % 移除摘要下方ACM引用格式区块
% \setcopyright{none}                   % 完全禁用版权声明
% \renewcommand\footnotetextcopyrightpermission[1]{} % 删除第一页底部会议版权脚注
% \pagestyle{plain}

% \begin{document}

% \settopmatter{printacmref=false}

% \clearpage
% \setcounter{page}{1}
% \title{Appendix}
\appendix

\makeatletter
\twocolumn[{
  \begin{center}
    \Huge \bfseries \@titlefont \textbf{Beyond Chain-of-Thought: Rewrite as a Universal Interface for
Generative Multimodal Embeddings}\\
  \end{center}
  \begin{center}
    \Huge \textit{Appendix}\\[1em]
  \end{center}
}]
\makeatother

\section{Data Construction Details}
\label{sec:data_construction}

% 中文分析：本节详细介绍了RIME模型训练数据的构建过程。数据构建是生成式嵌入模型的关键环节，
% 需要为查询端和目标端分别生成高质量的重写标注。我们采用了模态感知的提示模板，
% 针对文本、图像和视频三种模态设计了不同的重写策略。

This section provides a comprehensive overview of the data construction pipeline for training RIME.
The data construction process is critical for generative embedding models, as high-quality rewrite annotations for both queries and targets directly determine the model's retrieval performance.

\subsection{Training Data Composition}
\label{sec:data_composition}

% 中文分析：训练数据包含约150万个样本，来自多个公开数据集，涵盖图像、视频和视觉文档三种模态。
% 数据配比经过精心设计，确保模型在各模态上均衡学习。

The Rewrite SFT stage utilizes approximately \textbf{1.5 million} training samples drawn from the VLM2VEC-v2~\cite{meng2025vlm2vec} training datasets, spanning three modalities: images, videos, and visual documents. For image data, we incorporate VLM2Vec's image-related datasets including MSCOCO, ImageNet-1K, ChartQA, A-OKVQA, DocVQA, SUN397, Visual7W, N24News, VOC2007, HatefulMemes, VisualNews, InfographicsVQA, CIRR, VisDial, WebQA, NIGHTS, and OK-VQA, which cover diverse retrieval scenarios such as image-text matching, visual question answering, image classification, and composed image retrieval. For video data, we utilize LLaVA-Hound~\cite{zhang2025direct} for video-text retrieval and video question answering tasks, with three subsets including video QA, video retrieval, and caption retrieval to provide comprehensive coverage of video understanding scenarios. For visual document data, we employ ViDoRe~\cite{faysse2024colpali} and VisRAG~\cite{yu2024visrag} datasets for document-as-image retrieval tasks, which involve complex charts, tables, and paper figures requiring fine-grained text recognition and layout understanding.

For the Refine-RL stage, we curate a separate dataset of approximately \textbf{15K} samples that are uniformly sampled from three modalities (image, video, and visual document). Importantly, we ensure no overlap with the SFT training data to validate the model's generalization capability. This dataset serves as the exploration space for GRPO optimization, where the model learns to generate rewrites that maximize the process reward and so on.

\input{tables/10_datacom}

\subsection{Modality-Aware Rewrite Annotation}

\begin{figure*}[!t]
    \centering
    \includegraphics[width=\textwidth]{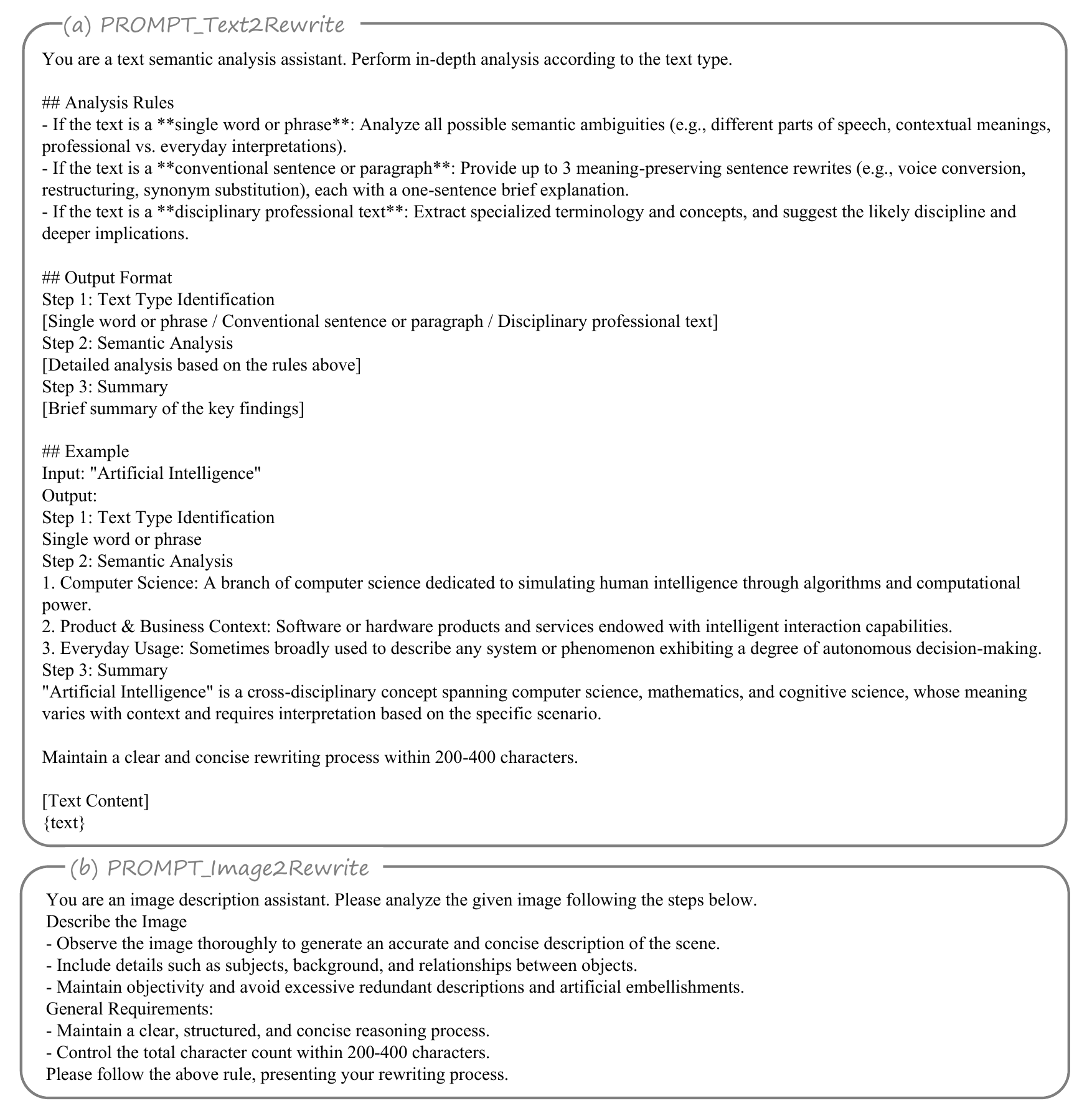}
    \caption{
    \textbf{Text-to-Rewrite (T2R) and Image-to-Rewrite (I2R)  Prompt Example.} 
    }
    \label{fig:T2R}
\end{figure*}

\begin{figure*}[!t]
    \centering
    \includegraphics[width=\textwidth]{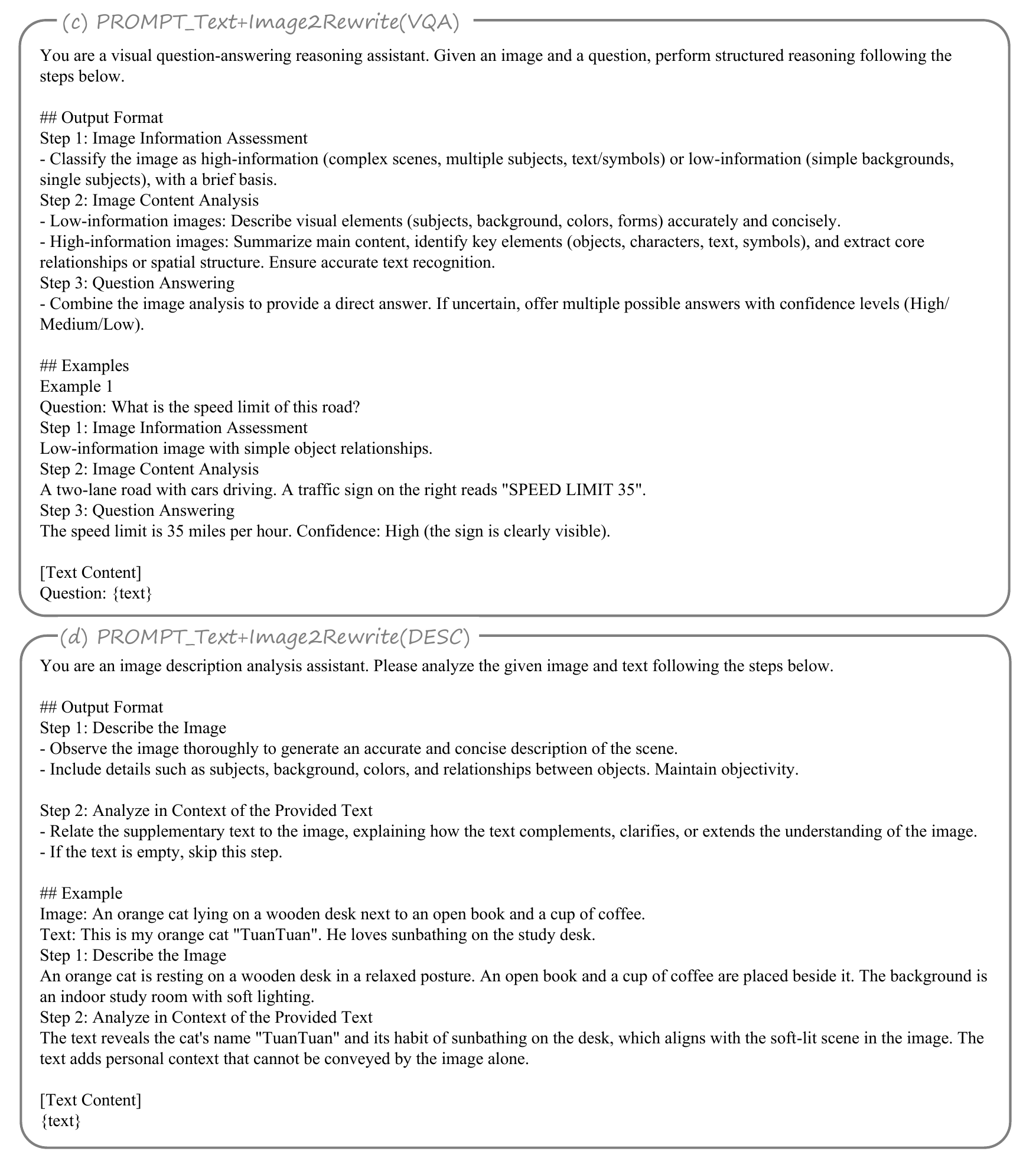}
    \caption{
    \textbf{Image-Text VQA to Rewrite (IT2R-VQA) and Image-Text Description to Rewrite (IT2R-DESC) Prompt Example.} 
    }
    \label{fig:VQAIT2R}
\end{figure*}

As shown in Figure~\ref{fig:T2R},~\ref{fig:VQAIT2R} and so on, we design modality-aware rewrite prompts that guide the model to generate retrieval-friendly descriptions~\cite{ma2023query,gao2022query}, categorized by input modality and task type. For pure text inputs, the text rewriting strategy performs semantic analysis including text type identification, ambiguity resolution, logical relationship extraction, and concise summarization. For image inputs, the image rewriting generates descriptive reconstructions that accurately capture subjects, backgrounds, colors, and object relationships while maintaining objectivity, with additional analysis of how accompanying text complements the visual content when image-text pairs are present. For visual question answering scenarios involving images, we employ a structured reasoning process that assesses image information density, performs OCR when necessary, and provides direct answers with confidence levels. For video inputs, the video rewriting strategy describes subjects, backgrounds, actions, scene transitions, and temporal progression, focusing on key frames and important event nodes while maintaining chronological order. For video-based visual question answering, we extend the image VQA approach to handle temporal dynamics, multiple events, and dialogue or text recognition across video frames. 

The rewrite annotations are generated using Qwen2-VL-72B-Instruct~\cite{wang2024qwen2} as the teacher model with an asynchronous batch processing strategy. The post-processing includes wrapping generated content in \texttt{<think>...</think>} tags, appending a standardized answer format, and filtering failed generations, resulting in final annotations that follow the format: \texttt{<think>\{rewrite\}</think>All can be embedded into <gen\_emb>}.

\input{tables/4_MR2}
\input{tables/8_ablation_lambda}

\section{Additional Experimental Results}
\label{sec:additional_results}

% 中文分析：本节提供了补充实验结果，包括超参数消融、各基准的详细子任务结果等，
% 以全面验证RIME的有效性和鲁棒性。

This section presents additional experimental results that complement the main paper, providing deeper insights into RIME's performance across various dimensions.

\subsection{Ablation on Hyperparameter $\lambda$}
\label{sec:ablation_lambda}

% 中文分析：超参数λ控制联合损失中重写生成损失与对比学习损失的权重平衡。
% 实验表明λ=1.0时性能最优，过小会导致重写质量下降，过大会削弱嵌入的判别能力。

The hyperparameter $\lambda$ in the joint loss function $\mathcal{L}_{\text{Joint}} = \lambda \cdot \mathcal{L}_{\text{Rewrite}} + \mathcal{L}_{\text{CM\_InfoNCE}}$ controls the trade-off between rewrite generation quality and retrieval performance.
As shown in Table~\ref{tab:ablation_lambda}, we conduct ablation studies on $\lambda$ using the MMEB-V2 benchmark. When $\lambda = 0.1$, the model achieves an overall score of 59.7, indicating that insufficient emphasis on rewrite generation leads to suboptimal semantic understanding. As $\lambda$ increases to 0.5, the performance improves to 61.1. The optimal performance of 64.1 is achieved at $\lambda = 1.0$, demonstrating that equal weighting of rewrite and contrastive objectives achieves the best balance. When $\lambda$ further increases to 1.5, the performance slightly decreases to 62.3, suggesting that over-emphasizing generation may weaken the discriminative capability of embeddings.

\subsection{Results on MR$^2$-Bench}
\label{sec:mr2_results}

% 中文分析：MR²-Bench是一个新兴的多模态推理检索基准，涵盖知识检索、视觉说明和视觉关系三大类别。
% RIME在该基准上展现了强大的推理能力，尤其在经济学和空间推理任务上表现突出。

The MR$^2$-Bench~\cite{zhou2025mr} is a recently proposed benchmark specifically designed to evaluate multimodal reasoning-intensive retrieval capabilities, comprising 12 sub-tasks across three categories: Multimodal Knowledge Retrieval (Biology, Cooking, Gardening, Physics, Chemistry, Earth Science), Visual Illustration (Economics, Mathematics, Nature), and Visual Relation (Spatial, Puzzle, Analogy). As shown in Table~\ref{tab:mr2}, RIME-7B achieves a comprehensive score of 28.6, outperforming UME-R1 (27.5) and approaching the closed-source Seed-1.6-embed (30.7). Notably, RIME demonstrates particularly strong performance on reasoning-intensive sub-tasks, with notable improvements observed in Economics (64.2 vs. 55.7 for UME-R1), Spatial reasoning (40.4 vs. 37.7), and Analogy (15.1 vs. 12.2), representing improvements of 8.5, 2.7, and 2.9 points respectively. Furthermore, the Puzzle sub-task remains challenging for all models with scores near zero, indicating that current multimodal embedding approaches still struggle with abstract visual reasoning tasks that require complex logical deduction. Overall, these results demonstrate that the rewrite-driven paradigm effectively activates internal knowledge for reasoning-intensive retrieval without relying on explicit chain-of-thought reasoning~\cite{lan2025ume,liu2025reasoning,zhu2025retrv}.

\subsection{Detailed Results on MMEB-V2/UVRB}
\label{sec:mmeb_detailed}

% 中文分析：MMEB-V2包含78个子任务，覆盖图像、视频和视觉文档三种模态。
% 我们提供各子任务的详细结果，展示RIME在不同检索场景下的性能表现。

To facilitate fair comparison and reproducible research for future work, we provide comprehensive per-task results on the MMEB-v2~\cite{jiang2024vlm2vec,meng2025vlm2vec} benchmark. Specifically, we present detailed evaluation results across all 78 sub-tasks spanning three modalities, including image tasks, video tasks, and visual document tasks, as summarized in Table~\ref{tab:detailed_score_part1} and~\ref{tab:detailed_score_part2}. These fine-grained results are intended to support further analysis and direct comparison in subsequent studies.
Based on the UVRB~\cite{guo2025towards} benchmark, which systematically evaluates video retrieval across 16 sub-tasks spanning three dimensions (Tasks, Domains, and Sub-domains), RIME-7B achieves the best overall performance with an average score of 55.6. Detailed per-task results are provided in Table~\ref{tab:main_by_datasets} and~\ref{tab:dataset_partition} to facilitate fair comparison and reproducible research for future work.

\section{Examples of Data Construction}
\label{sec:data_examples}

% 中文分析：本节提供了不同输入类型的数据构建示例，展示了模态感知重写的具体效果。
% 这些示例直观地说明了重写机制如何将多模态输入转换为检索友好的语义表示。

\begin{figure*}[!t]
    \centering
    \includegraphics[width=\textwidth]{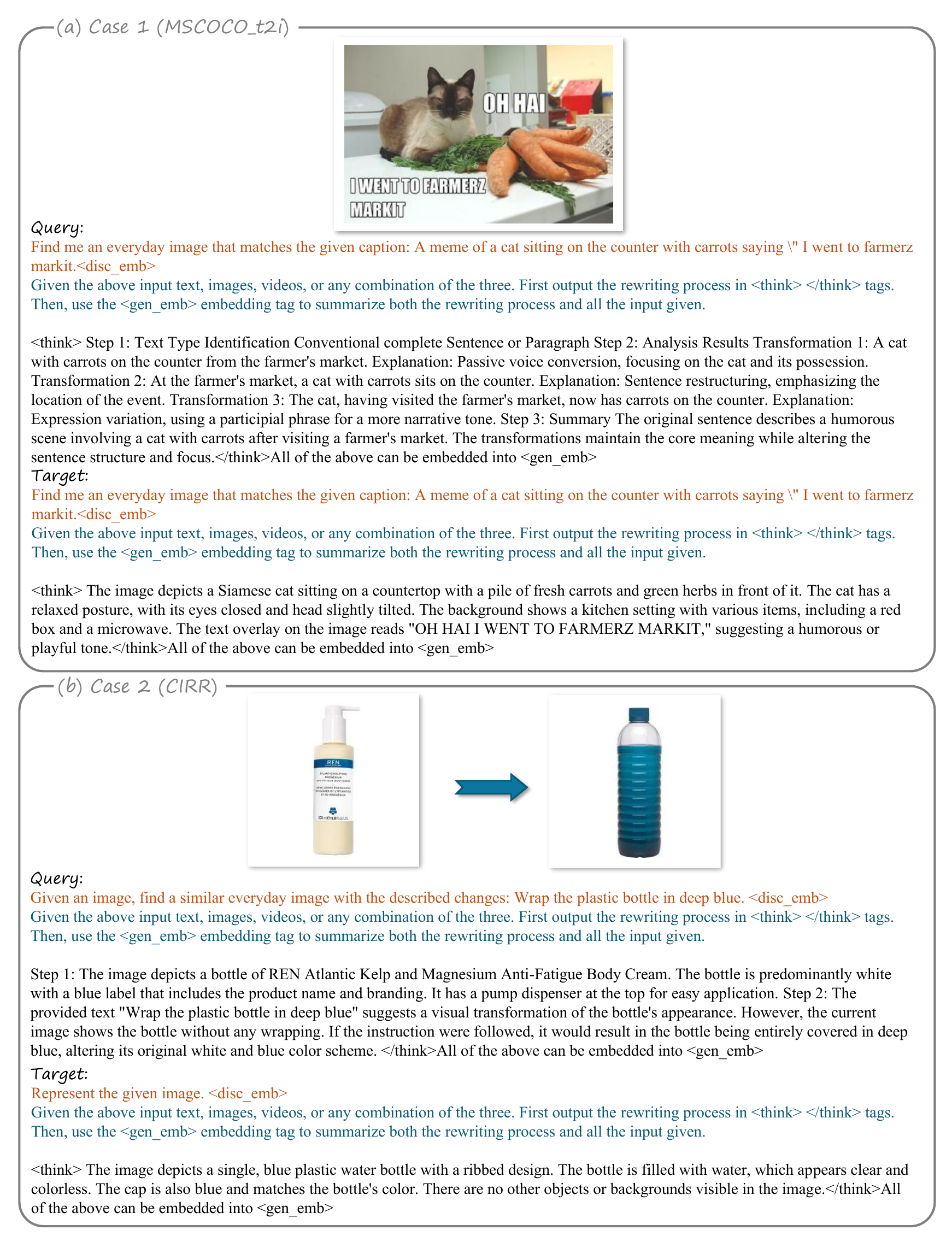}
    \caption{
    \textbf{Examples of Data Construction.} 
    }
    \label{fig:case1}
\end{figure*}

\begin{figure*}[!t]
    \centering
    \includegraphics[width=\textwidth]{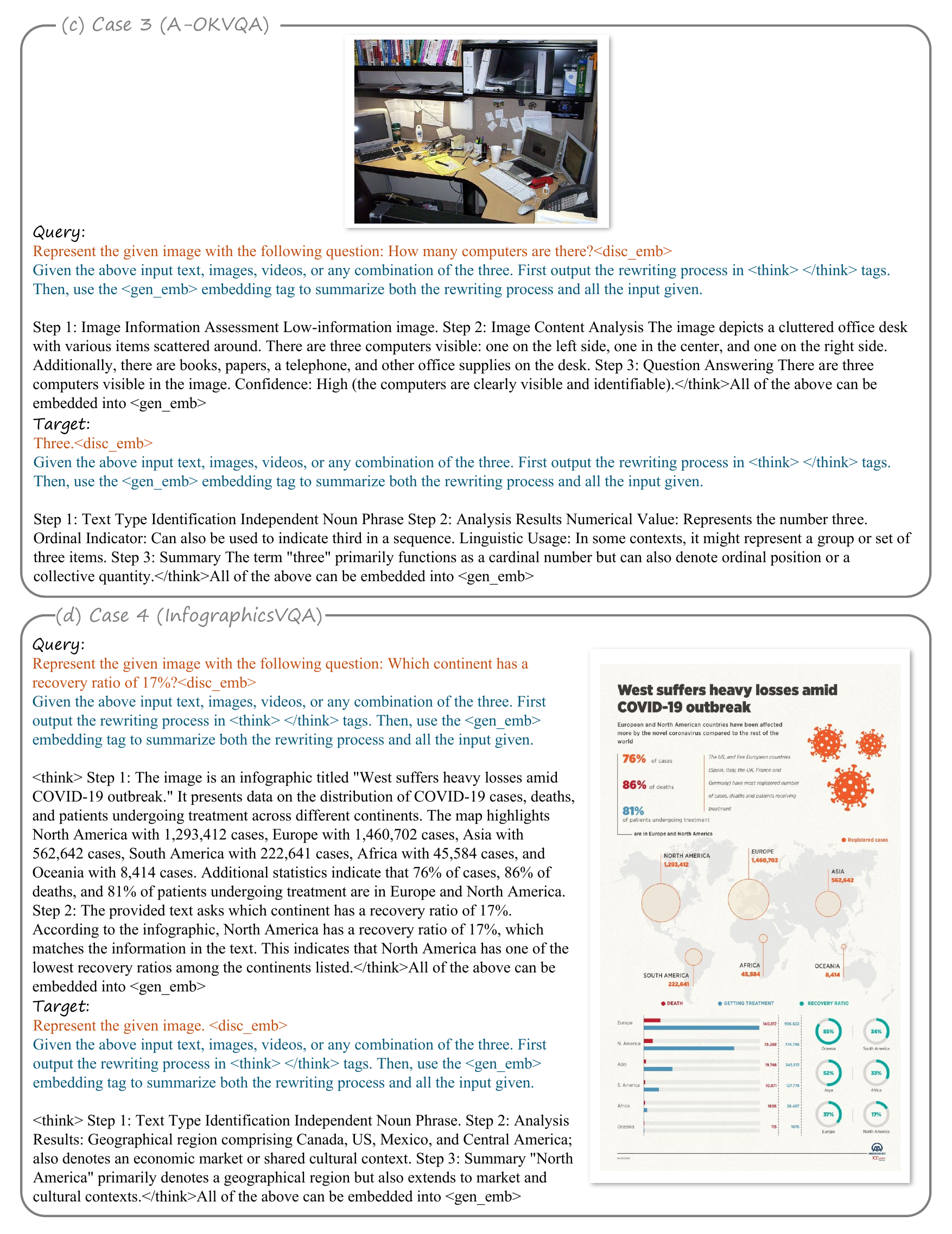}
    \caption{
    \textbf{Examples of Data Construction.} 
    }
    \label{fig:case2}
\end{figure*}

\begin{figure*}[!t]
    \centering
    \includegraphics[width=\textwidth]{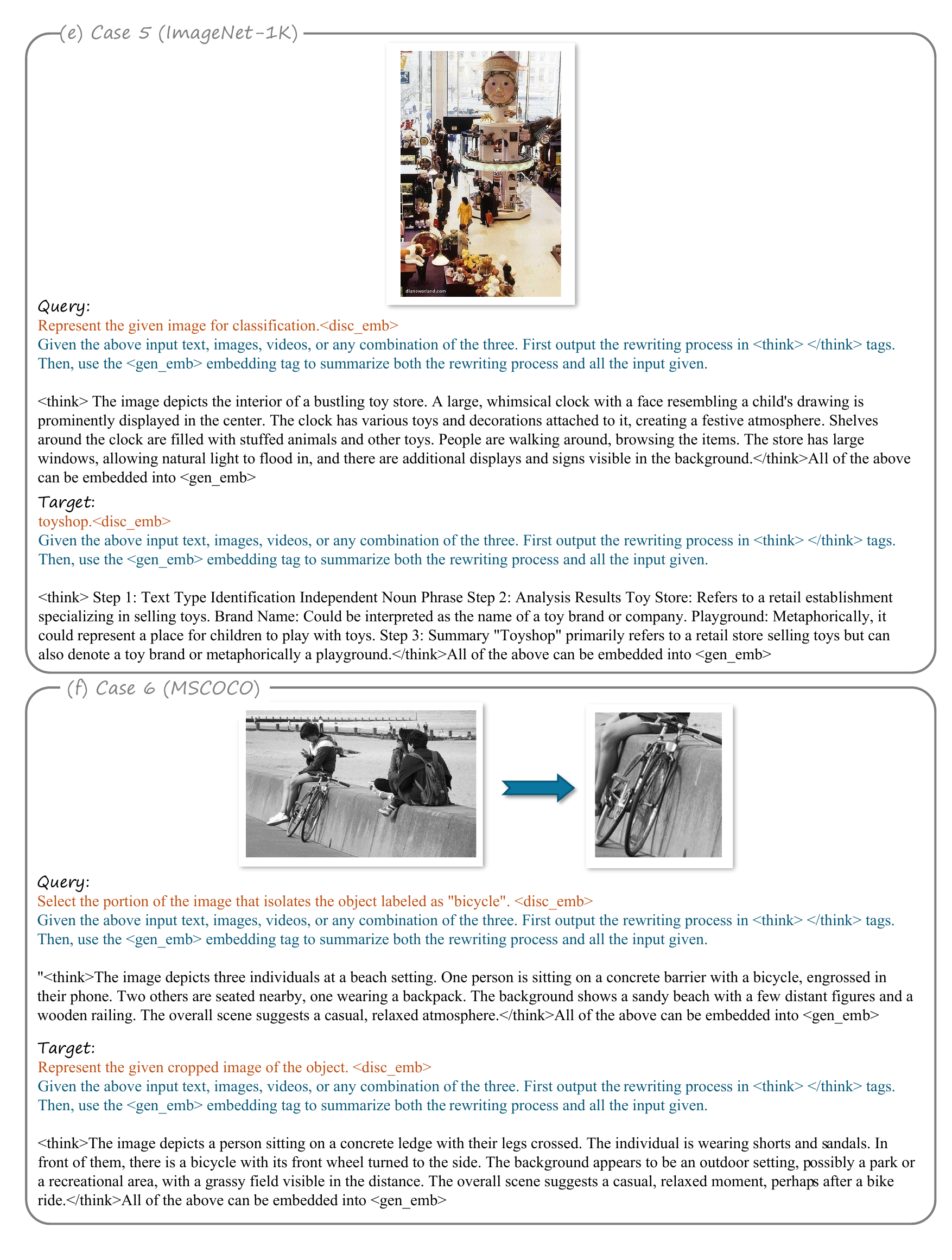}
    \caption{
    \textbf{Examples of Data Construction.} 
    }
    \label{fig:case3}
\end{figure*}

\begin{figure*}[!t]
    \centering
    \includegraphics[width=\textwidth]{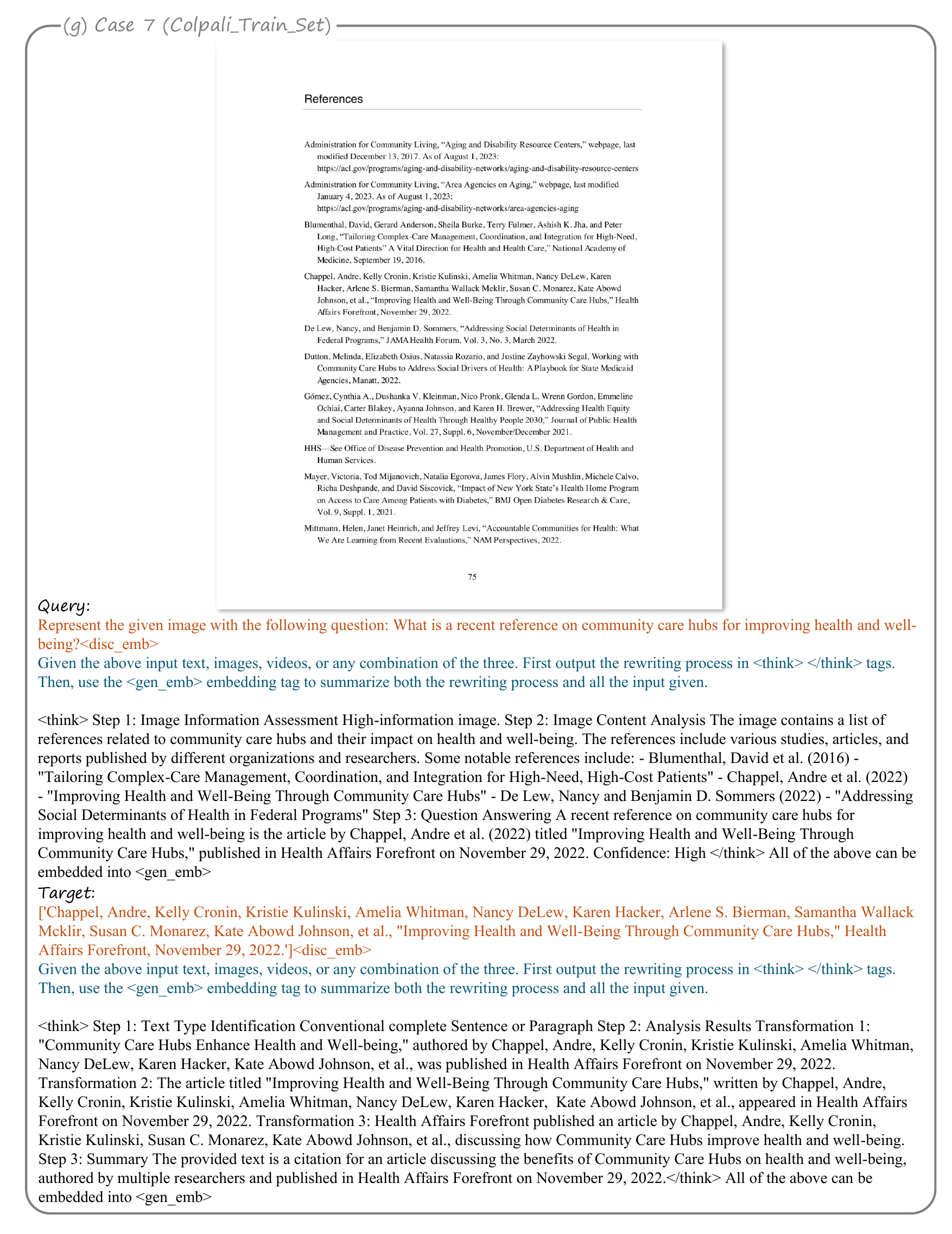}
    \caption{
    \textbf{Examples of Data Construction.} 
    }
    \label{fig:case4}
\end{figure*}

\begin{figure*}[!t]
    \centering
    \includegraphics[width=\textwidth]{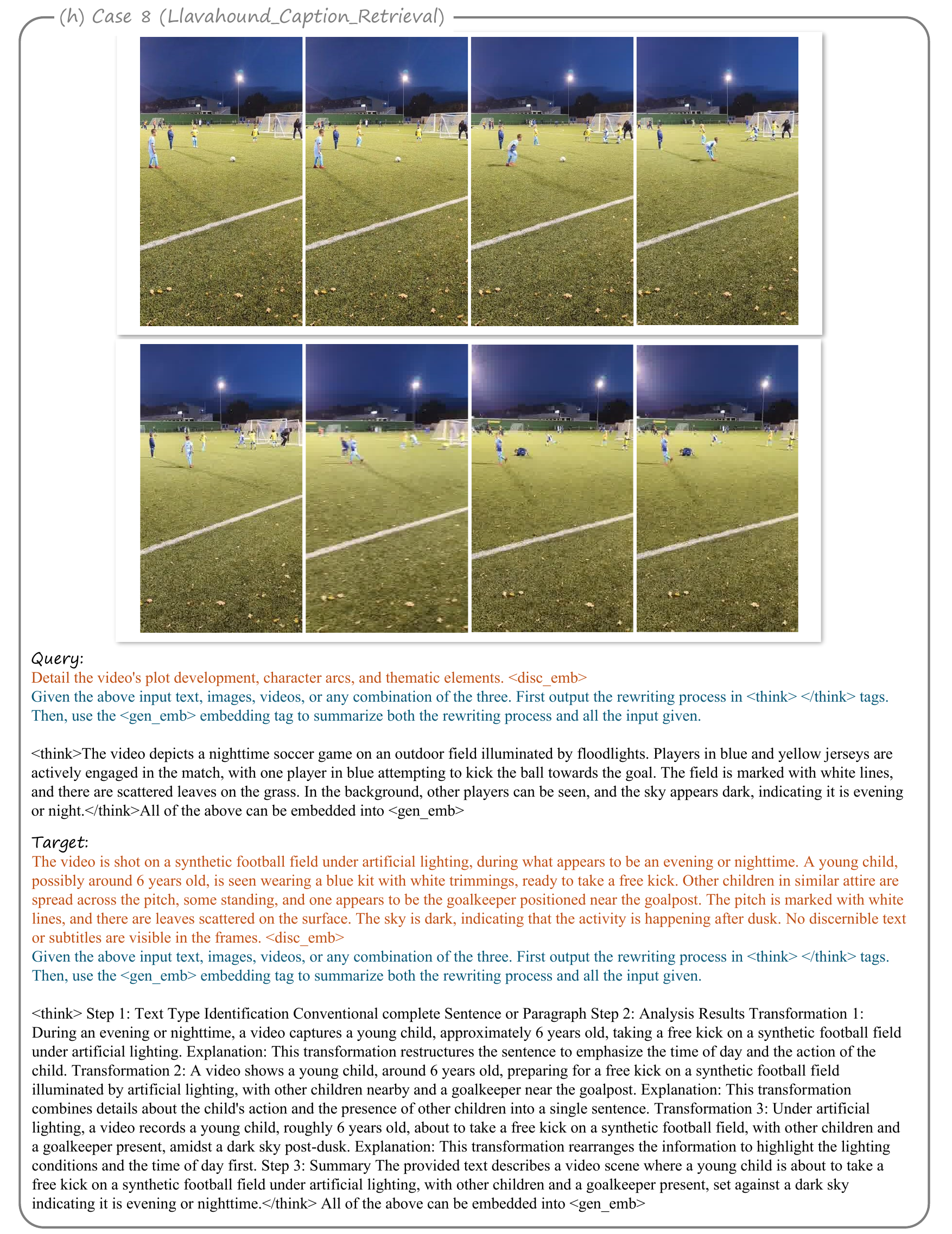}
    \caption{
    \textbf{Examples of Data Construction.} 
    }
    \label{fig:case5}
\end{figure*}

This section provides illustrative examples of the data construction process across different input modalities, demonstrating how the modality-aware rewriting transforms multimodal inputs into retrieval-friendly representations. As illustrated in Figures~\ref{fig:case1} -~\ref{fig:case5}, these examples demonstrate the concrete details of the rewriting process and are provided for reference.

\input{tables/11_mmeb}

\input{tables/12_urvb}
\input{tables/13_urvb_sub}

% \clearpage
% \balance
% \bibliographystyle{ACM-Reference-Format}
% \bibliography{papers/bibfile}

% \end{document}

%% file: tables/10_datacom.tex
\begin{table}[htbp]
    \centering
    \setlength{\tabcolsep}{7pt}
    \small
    \caption{Statistics of training data composition.}
    \label{tab:cot_filter_stats}
    \resizebox{\linewidth}{!}{
        \begin{tabular}{@{}ccccc@{}}
            \toprule
            \textbf{Dataset} & \textbf{Initial} & \textbf{Filtered} & \textbf{Ratio} & \textbf{Modality} \\
            \midrule
            \rowcolor[HTML]{E7E7F9}
            \multicolumn{5}{c}{\textit{Image-based (MMEB-train)}} \\
            A-OKVQA & 17,000 & 15,304 & 90.02\% & Text-Image → Text \\
            CIRR & 26,000 & 22,189 & 85.34\% & Text-Image → Text-Image \\
            ChartQA & 28,000 & 25,479 & 91.00\% & Text-Image → Text \\
            DocVQA & 40,000 & 37,049 & 92.62\% & Text-Image → Text \\
            HatefulMemes & 8,000 & 7,352 & 91.90\% & Text-Image → Text \\
            ImageNet-1K & 50,000 & 43,607 & 87.21\% & Text-Image → Text \\
            InfographicsVQA & 24,000 & 21,059 & 87.75\% & Text-Image → Text \\
            MSCOCO & 50,000 & 23,417 & 46.83\% & Text-Image → Text-Image \\
            MSCOCO-i2t & 50,000 & 44,630 & 89.26\% & Text-Image → Text \\
            MSCOCO-t2i & 50,000 & 43,229 & 86.46\% & Text → Text-Image \\
            N24News & 49,000 & 40,960 & 83.59\% & Text-Image → Text \\
            NIGHTS & 16,000 & 12,615 & 78.84\% & Text-Image → Text-Image \\
            OK-VQA & 9,000 & 8,022 & 89.13\% & Text-Image → Text \\
            SUN397 & 20,000 & 18,879 & 94.40\% & Text-Image → Text \\
            VOC2007 & 8,000 & 6,456 & 80.70\% & Text-Image → Text \\
            Visual7W & 50,000 & 44,382 & 88.76\% & Text-Image → Text \\
            VisDial & 50,000 & 39,537 & 79.07\% & Text → Text-Image \\
            VisualNews-i2t & 50,000 & 41,177 & 82.35\% & Text-Image → Text \\
            VisualNews-t2i & 50,000 & 39,823 & 79.65\% & Text → Text-Image \\
            WebQA & 17,000 & 12,978 & 76.34\% & Text → Text-Image \\
            
            \rowcolor[HTML]{E7E7F9}
            \multicolumn{5}{c}{\textit{Video-based (LLaVA-Hound)}} \\
            Caption Retrieval & 300,000 & 232,044 & 77.35\% & Video → Text \\
            Video QA & 300,000 & 207,829 & 69.28\% & Video-Text → Text \\
            Video Retrieval & 300,000 & 244,327 & 81.44\% & Text → Video \\
            
            \rowcolor[HTML]{E7E7F9}
            \multicolumn{5}{c}{\textit{Document-based}} \\
            ViDoRe & 100,000 & 90,415 & 90.42\% & Text-Image → Text \\
            VisRAG & 100,000 & 81,086 & 81.09\% & Text → Image \\
            
            \midrule
            \textbf{Image-based} & 662,000 & 548,144 & 82.80\% & Image-centric \\
            \textbf{Video-based} & 900,000 & 684,200 & 76.02\% & Video-centric \\
            \textbf{Document-based} & 200,000 & 171,501 & 85.75\% & Document-centric \\
            \textbf{Total} & 1,762,000 & 1,403,845 & 79.67\% & Multimodal \\
            \bottomrule
        \end{tabular}
    }
\end{table}

%% file: tables/4_MR2.tex
\begin{table*}[t]
\caption{Results on the \textbf{MR$^2$-Bench} benchmark. We report nDCG@10 for all sub-tasks: Biology (Bio.), Cooking (Cook.), Gardening (Gar.), Physics (Phy.), Chemistry (Chem.), Earth Science (Earth.), Economics (Econ.), Mathematics (Math.), Nature (Nat.), Spatial (Spa.), Puzzle (Puzz.), and Analogy (Ana.). All denotes the average score across 12 datasets.}
\label{tab:mr2}

\centering
\tiny
\setlength{\tabcolsep}{3.5pt}   
\renewcommand{\arraystretch}{1.05}

\resizebox{\textwidth}{!}{%
\begin{tabular}{llc cccccc ccc ccc c}
\toprule
\multirow{2}{*}{\textbf{Model}} 
& \multirow{2}{*}{\textbf{Backbone}}
& \multirow{2}{*}{\textbf{Size}}
& \multicolumn{6}{c}{\textbf{Multimodal Knowledge Retrieval}} 
& \multicolumn{3}{c}{\textbf{Visual Illustration}} 
& \multicolumn{3}{c}{\textbf{Visual Relation}} 
& \multirow{2}{*}{\textbf{All}} \\
\cmidrule(lr){4-9} \cmidrule(lr){10-12} \cmidrule(lr){13-15}
& & & Bio. & Cook. & Gar. & Phy. & Chem. & Earth. & Econ. & Math. & Nat. & Spa. & Puzz. & Ana. & \\
\midrule
\multicolumn{16}{c}{\emph{Closed-Source or w/ Additional Data}} \\
\midrule
Seed-1.6-embed & Seed1.6-flash & -- &
40.6 & 38.1 & 31.8 & 27.9 & 17.8 & 37.2 & 56.1 & 26.1 & 65.2 & 17.3 & 0.9 & 9.2 & 30.7 \\
\midrule
\multicolumn{16}{c}{\emph{Baseline Models}} \\
\midrule
CLIP & ViT-L/14 & 0.4B &
32.9 & 30.6 & 14.1 & 14.9 & 3.5 & 33.2 & 13.0 & 5.6 & 49.3 & 20.9 & 0.2 & 5.1 & 18.6 \\
BGE-VL & BGE-VL & 2B &
29.4 & 18.4 & 10.5 & 19.5 & 7.1 & 19.7 & 50.8 & 14.3 & 48.0 & 6.5 & 0.0 & 0.8 & 19.5 \\
GME & Qwen2-VL & 2B &
34.3 & 39.5 & 19.0 & 19.3 & 7.7 & 28.6 & 37.0 & 7.2 & 39.4 & 15.7 & 0.2 & 11.1 & 21.6 \\
VLM2Vec & Qwen2-VL & 2B &
39.4 & 39.4 & 19.9 & 20.3 & 9.0 & 35.7 & 51.4 & 14.2 & 35.1 & 13.9 & 0.6 & 5.9 & 23.7 \\
MM-Embed & NV-Embed & 8B &
\textbf{49.7} & \textbf{52.2} & 23.7 & \textbf{30.4} & \textbf{17.4} & \textbf{47.5} & 43.0 & 21.6 & 48.4 & 22.8 & 0.2 & 5.9 & \textbf{30.2} \\
UME-R1 & Qwen2-VL & 7B &
38.2 & 32.6 & 25.4 & 19.6 & 8.7 & 26.3 & 55.7 & 23.2 & 49.2 & 37.7 & 0.4 & 12.2 & 27.5 \\
\rowcolor[HTML]{E7E7F9}
\textbf{RIME (Ours)} & Qwen2-VL & 7B &
38.5 & 32.0 & \textbf{25.6} & 20.4 & 7.1 & 23.3 & \textbf{64.2} & \textbf{24.2} & \textbf{50} & \textbf{40.4} & \textbf{1.2} & \textbf{15.1} & 28.6 \\

\bottomrule
\end{tabular}%
}
\end{table*}

%% file: tables/8_ablation_lambda.tex
% ----------------- Ablation on Hyperparameter λ -----------------
\begin{table}[t]
\caption{Ablation study on the hyperparameter $\lambda$ in the joint loss of Rewrite SFT. We report the average scores on MMEB-V2 benchmark.}

\centering
\tiny
\setlength{\tabcolsep}{10pt}   
\renewcommand{\arraystretch}{0.95}
\resizebox{\columnwidth}{!}{%
\begin{tabular}{l cccc}
\toprule
\textbf{$\lambda$} & \textbf{Image} & \textbf{Video} & \textbf{VisDoc} & \textbf{All} \\
\midrule
0.1 & 64.8 & 38.2 & 68.5 & 59.7 \\
0.5 & 66.3 & 39.0 & 69.2 & 61.1 \\
\rowcolor[HTML]{E7E7F9}
1.0 & \textbf{69.1} & \textbf{43.7} & 71.4 & \textbf{64.1} \\
1.5 & 67.3 & 39.7 & \textbf{72.0} & 62.3 \\
\bottomrule
\end{tabular}
}

\label{tab:ablation_lambda}
\end{table}

%% file: tables/11_mmeb.tex
\definecolor{avgcolor}{RGB}{240,248,255}
\definecolor{icatcolor}{RGB}{255,245,235}
\definecolor{vcatcolor}{RGB}{240,255,240}
\definecolor{vdcatcolor}{RGB}{245,240,255}

\begin{table*}[htbp]
    \centering
    \caption{Detailed results of baselines and RIME on MMEB-v2 benchmark. Video and VisDoc results are shown on the next table. }
    \renewcommand{\arraystretch}{1.1}
    \resizebox{\textwidth}{!}{
    \begin{tabular}{l|cccccccccc}
        \toprule
        ~ & ColPali & GME-7B & VLM2Vec-7B & VLM2Vec-V2-2B & CAFe-7B & UME-R1-2B & UME-R1-7B & RIME-2B & RIME-7B \\ 
        \midrule
        \rowcolor{avgcolor}
        Avg - All (78 tasks) & 44.4 & 57.8 & 52.3 & 58.0 & 60.6 & 60.1 & 64.5 & 64.1 & 68.6 \\
        \midrule
        \rowcolor{avgcolor}
        Avg - Image (36 tasks, Hit@1) & 34.9 & 56.0 & 65.5 & 64.9 & 67.6 & 66.6 & 71.3 & 69.1 & 73.4 \\
        \rowcolor{avgcolor}
        Avg - Video (18 tasks, Hit@1) & 28.2 & 38.4 & 33.7 & 34.6 & 42.4 & 42.2 & 47.5 & 43.7 & 49.4 \\
        \rowcolor{avgcolor}
        Avg - Visdoc (24 tasks, NDCG@5) & 71.0 & 75.2 & 46.4 & 65.4 & 63.9 & 63.9 & 67.1 & 71.4 & 75.6 \\
        \midrule
        \rowcolor{icatcolor}
        I-CLS (10) & 40.3 & 57.7 & 62.7 & 62.9 & 63.6 & 64.8 & 67.1 & 67.9 & 70.3 \\
        \rowcolor{icatcolor}
        I-QA (10) & 11.5 & 34.7 & 56.9 & 56.3 & 61.7 & 62.8 & 69.2 & 64.4 & 71.7 \\
        \rowcolor{icatcolor}
        I-RET (12) & 48.1 & 71.2 & 69.4 & 69.5 & 69.1 & 67.6 & 71.9 & 69.8 & 73.2 \\
        \rowcolor{icatcolor}
        I-VG (4) & 40.3 & 59.3 & 82.2 & 77.3 & 87.6 & 77.2 & 84.9 & 82.1 & 86.3 \\
        \rowcolor{vcatcolor}
        V-CLS (5) & 26.7 & 37.4 & 39.1 & 39.3 & 35.8 & 44.3 & 48.6 & 48.0 & 52.6 \\
        \rowcolor{vcatcolor}
        V-QA (5) & 37.8 & 50.4 & 30.0 & 34.3 & 58.7 & 51.0 & 60.7 & 52.1 & 62.0 \\
        \rowcolor{vcatcolor}
        V-RET (5) & 21.6 & 28.4 & 29.0 & 28.8 & 34.4 & 32.9 & 38.2 & 33.6 & 38.4 \\
        \rowcolor{vcatcolor}
        V-MR (3) & 25.5 & 37.0 & 38.9 & 36.8 & 39.5 & 39.7 & 39.3 & 39.2 & 41.6 \\
        \rowcolor{vdcatcolor}
        VD-Vidore-V1 (10) & 83.6 & 89.4 & 56.9 & 75.7 & 70.7 & 72.4 & 75.7 & 76.4 & 80.9 \\
        \rowcolor{vdcatcolor}
        VD-Vidore-V2 (4) & 52.0 & 55.6 & 9.4 & 45.1 & 49.6 & 46.2 & 50.5 & 51.4 & 55.6 \\
        \rowcolor{vdcatcolor}
        VD-VisRAG (6) & 81.1 & 85.0 & 59.1 & 79.6 & 79.5 & 79.2 & 83.7 & 81.7 & 85.8 \\
        \rowcolor{vdcatcolor}
        VD-OOD (4) & 43.1 & 44.4 & 38.1 & 39.6 & 38.1 & 37.2 & 37.6 & 63.9 & 66.9 \\
        \midrule

        ImageNet-1K & 42.4 & 64.6 & 80.1 & 80.8 & 77.3 & 75.3 & 80.4 & 81.2 & 80.9 \\
        N24News & 25.5 & 50.5 & 79.7 & 72.9 & 83.2 & 81.1 & 82.3 & 80.0 & 82.7 \\
        HatefulMemes & 50.6 & 53.6 & 69.7 & 56.3 & 78.7 & 75.2 & 79.0 & 68.4 & 76.2 \\
        VOC2007 & 69.8 & 80.3 & 80.7 & 85.0 & 89.8 & 80.0 & 90.8 & 90.4 & 91.0 \\
        SUN397 & 56.1 & 69.5 & 77.4 & 71.0 & 79.9 & 79.4 & 80.3 & 80.1 & 80.6 \\
        Place365 & 27.5 & 39.1 & 37.4 & 35.9 & 45.0 & 42.6 & 46.8 & 45.3 & 45.5 \\
        ImageNet-A & 14.9 & 41.2 & 58.1 & 47.4 & 55.2 & 50.4 & 53.9 & 52.1 & 57.4 \\
        ImageNet-R & 64.6 & 83.9 & 73.9 & 89.3 & 88.0 & 88.7 & 90.1 & 89.9 & 89.9 \\
        ObjectNet & 45.6 & 69.0 & 40.1 & 65.2 & 22.5 & 52.0 & 42.3 & 66.1 & 72.7 \\
        Country211 & 6.0 & 24.8 & 29.8 & 25.2 & 16.7 & 23.4 & 25.0 & 25.5 & 26.4 \\
        OK-VQA & 9.4 & 33.2 & 56.8 & 51.5 & 67.3 & 62.4 & 71.7 & 65.8 & 74.1 \\
        A-OKVQA & 6.6 & 21.0 & 47.3 & 43.6 & 63.8 & 51.1 & 58.7 & 56.4 & 61.8 \\
        DocVQA & 11.3 & 41.4 & 89.7 & 90.1 & 79.2 & 92.2 & 93.8 & 93.4 & 94.4 \\
        InfographicsVQA & 5.0 & 20.3 & 60.0 & 58.8 & 53.3 & 67.7 & 79.2 & 63.6 & 79.1 \\
        ChartQA & 5.7 & 17.8 & 56.9 & 47.4 & 48.8 & 64.9 & 75.1 & 61.7 & 77.4 \\
        Visual7W & 6.1 & 22.2 & 52.7 & 52.9 & 52.5 & 54.1 & 55.2 & 54.8 & 54.9 \\
        ScienceQA & 16.3 & 28.0 & 38.5 & 38.2 & 65.4 & 42.7 & 53.7 & 45.7 & 59.0 \\
        VizWiz & 27.6 & 39.0 & 39.9 & 43.3 & 43.8 & 46.8 & 51.6 & 47.7 & 55.3 \\
        GQA & 8.3 & 76.9 & 55.1 & 64.9 & 65.7 & 67.3 & 69.3 & 73.1 & 73.6 \\
        TextVQA & 18.8 & 46.8 & 71.6 & 72.2 & 76.8 & 78.6 & 83.5 & 81.3 & 87.0 \\
        VisDial & 41.2 & 60.8 & 81.9 & 82.7 & 82.7 & 76.6 & 80.7 & 80.7 & 82.6 \\
        CIRR & 8.2 & 54.9 & 51.1 & 57.5 & 60.4 & 53.7 & 55.3 & 57.0 & 60.0 \\
        VisualNews\_t2i & 50.1 & 79.7 & 80.5 & 74.5 & 69.5 & 71.7 & 76.8 & 72.4 & 79.8 \\
        VisualNews\_i2t & 47.6 & 83.6 & 81.2 & 78.2 & 79.4 & 74.2 & 82.0 & 78.2 & 83.5 \\
        MSCOCO\_t2i & 59.2 & 71.2 & 77.2 & 75.3 & 75.4 & 75.1 & 78.3 & 76.1 & 77.8 \\
        MSCOCO\_i2t & 49.9 & 57.7 & 73.9 & 71.4 & 73.1 & 68.9 & 71.4 & 70.0 & 72.6 \\
        NIGHTS & 65.5 & 67.6 & 67.6 & 68.6 & 66.7 & 67.2 & 68.1 & 67.9 & 68.6 \\
        WebQA & 53.8 & 91.4 & 88.3 & 90.6 & 89.3 & 90.0 & 90.9 & 90.7 & 90.7 \\
        FashionIQ & 5.9 & 37.8 & 17.1 & 19.5 & 39.0 & 17.1 & 23.4 & 19.8 & 26.0 \\
        Wiki-SS-NQ & 80.5 & 78.2 & 62.3 & 66.9 & 61.2 & 62.0 & 72.5 & 68.9 & 76.3 \\
        OVEN & 50.0 & 75.1 & 66.5 & 64.3 & 60.8 & 66.9 & 71.4 & 67.5 & 68.6 \\
        EDIS & 64.7 & 96.0 & 85.7 & 84.1 & 71.3 & 88.0 & 92.0 & 88.9 & 91.6 \\
        MSCOCO & 36.7 & 31.4 & 75.7 & 67.1 & 84.7 & 69.5 & 72.7 & 69.0 & 72.0 \\
        RefCOCO & 64.5 & 60.9 & 87.6 & 87.1 & 89.4 & 83.3 & 91.4 & 88.9 & 91.7 \\
        RefCOCO-Matching & 3.9 & 78.4 & 84.6 & 85.8 & 83.0 & 84.4 & 91.1 & 90.1 & 93.7 \\
        Visual7W-Pointing & 56.1 & 66.5 & 81.0 & 69.2 & 93.2 & 71.5 & 84.2 & 80.5 & 87.9 \\
        \bottomrule
    \end{tabular}
    }
\label{tab:detailed_score_part1}
\end{table*}

\begin{table*}[htbp]
    \centering
    \caption{Detailed results of baselines and RIME on Video and Visual Doc of MMEB-v2 benchmark.}
    \renewcommand{\arraystretch}{1.1}
    \resizebox{\textwidth}{!}{
    \begin{tabular}{l|cccccccccc}
        \toprule
        ~ & ColPali & GME-7B & VLM2Vec-7B & VLM2Vec-V2-2B & CAFe-7B & UME-R1-2B & UME-R1-7B & RIME-2B & RIME-7B \\ 
        \midrule

        K700 & 23.4 & 39.7 & 35.5 & 38.0 & 40.1 & 35.8 & 42.8 & 47.7 & 55.0 \\
        SmthSmthV2 & 25.1 & 30.6 & 32.1 & 42.8 & 35.8 & 44.1 & 50.4 & 48.5 & 55.1 \\
        HMDB51 & 24.8 & 47.9 & 42.2 & 40.9 & 46.9 & 54.4 & 58.3 & 56.7 & 58.9 \\
        UCF101 & 49.4 & 54.7 & 61.8 & 60.0 & 39.6 & 67.2 & 70.0 & 66.4 & 68.5 \\
        Breakfast & 10.9 & 14.3 & 23.8 & 14.8 & 16.6 & 20.1 & 21.5 & 20.6 & 25.4 \\
        MVBench & 33.7 & 46.6 & 28.5 & 33.7 & 48.9 & 49.9 & 58.2 & 49.1 & 59.9 \\
        Video-MME & 30.6 & 39.2 & 27.8 & 30.7 & 46.0 & 41.7 & 47.3 & 41.6 & 49.6 \\
        NExTQA & 35.2 & 53.6 & 20.3 & 20.9 & 62.4 & 59.9 & 69.6 & 58.9 & 69.7 \\
        EgoSchema & 38.4 & 46.8 & 21.8 & 34.0 & 60.0 & 45.4 & 52.4 & 40.4 & 55.6 \\
        ActivityNetQA & 51.3 & 65.6 & 51.4 & 52.3 & 76.0 & 57.8 & 76.0 & 70.5 & 75.1 \\
        DiDeMo & 22.8 & 26.4 & 29.3 & 30.4 & 37.8 & 32.4 & 40.0 & 33.8 & 38.4 \\
        MSR-VTT & 17.6 & 31.8 & 34.5 & 28.3 & 36.5 & 34.3 & 38.9 & 36.4 & 41.5 \\
        MSVD & 45.4 & 49.7 & 46.7 & 48.1 & 56.4 & 55.4 & 60.8 & 56.6 & 59.4 \\
        VATEX & 16.7 & 24.9 & 25.5 & 26.5 & 32.0 & 29.9 & 32.6 & 28.1 & 32.7 \\
        YouCook2 & 5.3 & 9.1 & 9.0 & 10.6 & 9.5 & 12.7 & 18.5 & 13.3 & 19.9 \\
        QVHighlight & 19.9 & 59.5 & 57.7 & 49.4 & 58.4 & 57.5 & 54.9 & 55.1 & 56.9 \\
        Charades-STA & 29.0 & 14.0 & 19.8 & 20.2 & 18.7 & 20.4 & 21.9 & 19.4 & 21.6 \\
        MomentSeeker & 27.6 & 37.4 & 39.3 & 40.8 & 41.4 & 41.2 & 41.1 & 43.2 & 46.2 \\
        \midrule

        ViDoRe\_arxivqa & 81.7 & 86.9 & 60.2 & 80.6 & 73.3 & 73.9 & 73.6 & 82.1 & 84.1 \\
        ViDoRe\_docvqa & 56.6 & 57.5 & 34.7 & 44.9 & 38.3 & 37.9 & 41.1 & 45.4 & 46.9 \\
        ViDoRe\_infovqa & 84.9 & 91.6 & 70.4 & 83.7 & 80.6 & 76.2 & 80.8 & 81.2 & 85.4 \\
        ViDoRe\_tabfquad & 86.9 & 94.6 & 78.2 & 89.2 & 80.7 & 86.1 & 90.2 & 85.5 & 95.3 \\
        ViDoRe\_tatdqa & 70.9 & 74.1 & 27.6 & 43.8 & 37.8 & 40.6 & 46.7 & 48.9 & 52.6 \\
        ViDoRe\_shiftproject & 75.1 & 96.8 & 38.6 & 60.8 & 52.0 & 66.8 & 65.0 & 69.9 & 73.7 \\
        ViDoRe\_artificial\_intelligence & 95.7 & 99.6 & 67.7 & 88.5 & 86.0 & 85.9 & 89.5 & 89.8 & 96.1 \\
        ViDoRe\_energy & 94.7 & 95.3 & 60.4 & 86.5 & 84.8 & 83.3 & 85.7 & 86.2 & 88.5 \\
        ViDoRe\_government\_reports & 93.6 & 98.8 & 61.8 & 85.0 & 85.0 & 82.6 & 89.8 & 86.4 & 91.7 \\
        ViDoRe\_healthcare\_industry & 95.9 & 99.3 & 69.9 & 92.2 & 88.4 & 90.8 & 94.3 & 88.3 & 94.9 \\
        ViDoRe\_esg\_reports\_human\_labeled\_v2 & 51.3 & 63.4 & 6.8 & 45.6 & 50.7 & 50.2 & 50.4 & 56.5 & 60.2 \\
        ViDoRe\_biomedical\_lectures\_v2\_multilingual & 54.7 & 49.5 & 5.1 & 44.3 & 50.9 & 46.2 & 50.7 & 47.6 & 51.5 \\
        ViDoRe\_economics\_reports\_v2\_multilingual & 49.0 & 54.2 & 13.9 & 43.0 & 54.3 & 45.7 & 57.8 & 47.0 & 59.2 \\
        ViDoRe\_esg\_reports\_v2\_multilingual & 52.9 & 55.4 & 11.9 & 46.6 & 42.3 & 42.7 & 43.2 & 54.3 & 51.6 \\
        VisRAG\_ArxivQA & 80.9 & 87.4 & 52.6 & 76.9 & 74.0 & 74.3 & 80.5 & 79.4 & 84.0 \\
        VisRAG\_ChartQA & 72.3 & 86.1 & 57.7 & 83.7 & 82.7 & 86.0 & 85.0 & 83.3 & 85.4 \\
        VisRAG\_MP-DocVQA & 82.0 & 89.7 & 60.6 & 88.1 & 75.1 & 75.6 & 83.4 & 82.3 & 87.4 \\
        VisRAG\_SlideVQA & 85.1 & 92.6 & 54.7 & 84.1 & 87.6 & 87.1 & 91.5 & 91.0 & 94.1 \\
        VisRAG\_InfoVQA & 83.5 & 88.6 & 66.0 & 82.3 & 87.9 & 84.4 & 89.2 & 82.4 & 91.8 \\
        VisRAG\_PlotQA & 79.3 & 76.5 & 62.7 & 75.9 & 69.4 & 68.0 & 72.7 & 71.6 & 72.1 \\
        ViDoSeek-page & 38.1 & 32.6 & 16.3 & 29.1 & 22.5 & 21.2 & 21.3 & 80.7 & 85.6 \\
        ViDoSeek-doc & 87.5 & 90.3 & 69.4 & 79.0 & 73.8 & 75.9 & 75.3 & 79.7 & 80.9 \\
        MMLongBench-page & 27.1 & 36.9 & 0.4 & 15.8 & 13.3 & 11.9 & 12.3 & 48.3 & 52.4 \\
        MMLongBench-doc & 80.4 & 85.2 & 28.8 & 63.0 & 42.6 & 39.7 & 41.3 & 46.9 & 48.8 \\
        \bottomrule
    \end{tabular}
    }
\label{tab:detailed_score_part2}
\end{table*}

%% file: tables/12_urvb.tex
\begin{table*}[htbp]
\centering
\setlength{\tabcolsep}{9.5pt}  
\caption{Performance of video retrieval on UVRB datasets: AVG values represent the average across 16 datasets.}
\label{tab:main_by_datasets}
\renewcommand{\arraystretch}{0.95}
\resizebox{\textwidth}{!}{
\begin{tabular}{l|c|*{8}{c}}
\toprule
\textbf{Model} & \textbf{AVG} & \textbf{MSRVTT} & \textbf{DiDeMo} & \textbf{CRB-G} & \textbf{CRB-S} & \textbf{VDC-O} & \textbf{CRB-T} & \textbf{CMRB} & \textbf{DREAM-E} \\
& & R@1 & R@1 & R@1 & R@1 & R@1 & R@1 & R@10 & R@1 \\
\midrule
CLIP4Clip & 39.0 & 33.3 & 29.7 & 51.1 & 49.7 & 62.0 & 28.9 & 28.0 & 19.1 \\
ViCLIP & 35.2 & 38.6 & 30.6 & 44.7 & 43.7 & 53.0 & 34.9 & 22.9 & 23.5 \\
VideoCLIP-XL & 49.1 & 44.3 & 40.3 & 82.8 & 83.9 & 73.5 & 48.7 & 27.4 & 26.3 \\
LanguageBind & 48.7 & 47.9 & 42.1 & 71.6 & 68.7 & 75.9 & 46.6 & 29.0 & 28.0 \\
InternVideo2-1B & 40.4 & 44.9 & 40.4 & 58.6 & 56.8 & 64.4 & 47.0 & 35.5 & 24.2 \\
InternVideo2-6B & 42.7 & 48.5 & 41.8 & 60.8 & 61.2 & 65.0 & 45.5 & 34.6 & 27.1 \\
GME-2B & 48.8 & 39.0 & 30.3 & 69.0 & 71.8 & 71.5 & 40.0 & 29.8 & 24.0 \\
Unite-2B & 48.0 & 36.7 & 29.8 & 69.9 & 72.3 & 72.7 & 40.9 & 28.4 & 22.3 \\
VLM2Vec-V2 & 50.8 & 33.0 & 29.9 & 82.8 & 84.3 & 77.5 & 41.0 & 28.6 & 22.8 \\
BGE-VL & 44.3 & 33.7 & 31.8 & 69.0 & 68.8 & 63.9 & 35.9 & 22.5 & 21.2 \\
UniME-7B & 52.1 & 35.1 & 33.5 & 81.5 & 82.7 & 74.3 & 47.6 & 31.7 & 29.3 \\
B3-7B & 51.1 & 28.2 & 35.0 & 81.5 & 82.5 & 76.8 & 41.5 & 31.2 & 21.6 \\
GME-7B & 53.0 & 43.6 & 37.7 & 74.0 & 76.7 & 73.1 & 44.2 & 30.4 & 27.4 \\
Unite-7B & 53.8 & 43.9 & 38.6 & 79.8 & 80.4 & 75.3 & 47.2 & 35.1 & 27.9 \\
GVE-3B & 54.4 & 43.1 & 37.6 & 85.0 & 84.6 & 78.6 & 49.6 & 36.3 & 28.0 \\
GVE-7B & 57.3 & 46.4 & 43.3 & 86.5 & 84.7 & 79.4 & 53.9 & 39.8 & 30.2 \\
\midrule
\rowcolor[HTML]{E7E7F9}
RIME-2B (Ours) & 52.3 & 39.2 & 33.5 & 87.9 & 85.9 & 80.5 & 44.6 & 29.3 & 27.8 \\
\rowcolor[HTML]{E7E7F9}
RIME-7B (Ours) & 55.6 & 41.6 & 40.0 & 88.3 & 86.7 & 82.5 & 49.6 & 37.2 & 31.8 \\
\bottomrule
\end{tabular}
}
\vspace{5pt}

\resizebox{\textwidth}{!}{
\begin{tabular}{l|*{8}{c}}
\toprule
\textbf{Model} & \textbf{LoVR-TH} & \textbf{PEV-K} & \textbf{LoVR-V} & \textbf{VDC-D} & \textbf{MS-TI} & \textbf{MS-TV} & \textbf{MSRVTT-I2V} & \textbf{LoVR-C2V} \\
& R@10 & R@1 & R@1 & R@1 & P@1 & P@1 & R@1 & R@1 \\
\midrule
CLIP4Clip & 33.8 & 17.9 & 36.0 & 56.6 & 17.3 & 18.3 & 92.4 & 50.3 \\
ViCLIP & 20.2 & 7.5 & 23.0 & 39.5 & 28.3 & 24.3 & 84.6 & 43.3 \\
VideoCLIP-XL & 43.9 & 22.9 & 38.0 & 82.0 & 23.0 & 22.3 & 86.1 & 40.3 \\
LanguageBind & 42.5 & 30.3 & 54.0 & 67.9 & 22.8 & 23.3 & 82.7 & 46.3 \\
InternVideo2-1B & 29.8 & 2.6 & 28.0 & 48.5 & 26.5 & 23.0 & 79.4 & 36.8 \\
InternVideo2-6B & 30.2 & 8.6 & 33.0 & 51.6 & 23.5 & 20.5 & 86.8 & 45.2 \\
GME-2B & 44.6 & 35.4 & 53.0 & 83.9 & 35.0 & 34.0 & 82.7 & 36.6 \\
Unite-2B & 44.5 & 35.5 & 57.0 & 79.2 & 25.0 & 23.3 & 86.3 & 44.5 \\
VLM2Vec-V2 & 49.2 & 32.4 & 61.0 & 91.3 & 27.5 & 25.0 & 84.1 & 38.5 \\
BGE-VL & 38.7 & 18.4 & 55.0 & 72.2 & 30.3 & 23.3 & 77.9 & 46.5 \\
UniME-7B & 50.4 & 32.3 & 48.0 & 84.7 & 31.0 & 30.5 & 86.7 & 53.7 \\
B3-7B & 46.2 & 38.7 & 59.0 & 85.3 & 27.5 & 26.5 & 88.4 & 47.1 \\
GME-7B & 52.3 & 39.6 & 71.0 & 86.5 & 34.8 & 33.3 & 86.0 & 37.0 \\
Unite-7B & 55.5 & 44.0 & 62.0 & 87.1 & 27.8 & 23.0 & 88.3 & 44.8 \\
GVE-3B & 52.2 & 33.0 & 61.0 & 91.8 & 34.0 & 26.8 & 89.1 & 40.3 \\
GVE-7B & 54.2 & 41.3 & 68.0 & 94.8 & 34.3 & 28.0 & 89.9 & 41.5 \\
\midrule
\rowcolor[HTML]{E7E7F9}
RIME-2B (Ours) & 51.2 & 30.8 & 67.0 & 91.7 & 18.5 & 16.4 & 86.7 & 45.8 \\
\rowcolor[HTML]{E7E7F9}
RIME-7B (Ours) & 58.4 & 35.8 & 71.0 & 93.1 & 18.5 & 18.8 & 87.7 & 48.1 \\
\bottomrule
\end{tabular}
}
\end{table*}

%% file: tables/13_urvb_sub.tex
\begin{table*}[htbp]
  \centering
\scriptsize
\caption{Detailed Partition of Datasets in the UVRB Across Tasks, Domains, and Sub-domains}
\label{tab:dataset_partition}
\renewcommand{\arraystretch}{0.99}
\resizebox{\textwidth}{!}{
  \begin{tabular}{@{}p{2.5cm}@{}p{10cm}@{}}
    \toprule
    \textbf{Partition} & \textbf{Content} \\
    \midrule
    $\mathcal{D}_{\text{Txt.}}$ & $\{\text{MSRVTT}, \text{DiDeMo}, \text{CRB-G}, \text{CRB-S}, \text{VDC-O}, \text{CRB-T}, \text{CMRB},$ \\
    & $\quad \text{DREAM-E}, \text{LoVR-TH}, \text{PEV-K}, \text{LoVR-V}, \text{VDC-D}\}$ \\
    $\mathcal{D}_{\text{Cmp.}}$ & $\{\text{MS-TI}, \text{MS-TV}\}$ \\
    $\mathcal{D}_{\text{Vis.}}$ & $\{\text{MSRVTT-I2V}, \text{LoVR-C2V}\}$ \\
    $\mathcal{D}_{\text{Coarse-G.}}$  & $\{\text{MSRVTT}, \text{DiDeMo}, \text{CRB-G}\}$ \\
    $\mathcal{D}_{\text{Fine-G.}}$  & $\{\text{CRB-S}, \text{VDC-O}, \text{CRB-T}, \text{CMRB}, \text{DREAM-E}, \text{LoVR-TH}, \text{PEV-K}\}$ \\
    $\mathcal{D}_{\text{Long-Ctx.}}$  & $\{\text{LoVR-V}, \text{VDC-D}\}$ \\
    $\mathcal{D}_{\text{Spa.}}$   & $\{\text{CRB-S}, \text{VDC-O}\}$ \\
    $\mathcal{D}_{\text{Temp.}}$   & $\{\text{CRB-T}, \text{CMRB}\}$ \\
    $\mathcal{D}_{\text{PR.}}$  & $\{\text{DREAM-E}, \text{LoVR-TH}, \text{PEV-K}\}$ \\
    \bottomrule
  \end{tabular}
}
\end{table*}